%% file: emnlp2023.tex
\pdfoutput=1

\documentclass[11pt]{article}

\usepackage[]{EMNLP2023}

\usepackage{times}
\usepackage{latexsym}

\usepackage[T1]{fontenc}

\usepackage[utf8]{inputenc}

\usepackage{microtype}

\usepackage{inconsolata}

\usepackage{times}
\usepackage{latexsym}
\usepackage{graphics}
\usepackage{graphicx}
\usepackage{color}
\usepackage{colortbl}
\usepackage{amssymb}
\usepackage{amsmath}
\usepackage{bbm}
\usepackage{booktabs}
\usepackage{microtype}
\usepackage{multirow}
\usepackage{footmisc}
\usepackage{soul}
\usepackage{arydshln}

\usepackage{hyperref}
\usepackage{xspace}

\usepackage{algorithm}
\usepackage{algorithmic}

\usepackage{color}
\usepackage{url}

\newcommand{\eg}{\textit{e.g.}}

\usepackage{wasysym}
\definecolor{DarkGreen}{RGB}{0, 100, 0}
\definecolor{DarkBlue}{RGB}{0,0,139}

\title{Schema-adaptable Knowledge Graph Construction}

\author{
Hongbin Ye\textsuperscript{\rm 1,2,4}, 
Honghao Gui\textsuperscript{\rm 1,4},
Xin Xu\textsuperscript{\rm 1,4},
Xi Chen\textsuperscript{\rm 3},
Huajun Chen\textsuperscript{\rm 1,4},
Ningyu Zhang\textsuperscript{\rm 1,4 \thanks{\quad Corresponding author.}} \\
\textsuperscript{\rm 1} Zhejiang University
\textsuperscript{\rm 2} Zhejiang Lab 
\textsuperscript{\rm 3} Platform and Content Group, Tencent \\
\textsuperscript{\rm 4} Zhejiang University - Ant Group Joint Laboratory of Knowledge Graph\\
\texttt{yehongbin@zhejianglab.com},
jasonxchen@tencent.com,
\\ \texttt{\{guihonghao,xxucs,huajunsir,zhangningyu\}@zju.edu.cn} \\
}

\begin{document}
\maketitle
\begin{abstract}
Conventional Knowledge Graph Construction (KGC) approaches typically follow the static information extraction paradigm with a closed set of pre-defined \emph{schema}. As a result, such approaches fall short when applied to dynamic scenarios or domains, whereas a new type of knowledge emerges. This necessitates a system that can handle evolving schema automatically to extract information for KGC. To address this need, we propose a new task called schema-adaptable KGC, which aims to continually extract entity, relation, and event based on a dynamically changing schema graph without re-training. We first split and convert existing datasets based on three principles to build a benchmark, i.e., horizontal schema expansion, vertical schema expansion, and hybrid schema expansion; then investigate the schema-adaptable performance of several well-known approaches such as Text2Event, TANL, UIE and GPT-3.5. We further propose a simple yet effective baseline dubbed \textsc{AdaKGC}, which contains schema-enriched prefix instructor and schema-conditioned dynamic decoding to better handle evolving schema. Comprehensive experimental results illustrate that \textsc{AdaKGC} can outperform baselines but still have room for improvement. We hope the proposed work can deliver benefits to the community\footnote{Code and datasets available at \url{https://github.com/zjunlp/AdaKGC}.}.
\end{abstract}

\section{Introduction}

\begin{figure}
    \centering
    \resizebox{.49\textwidth}{!}{
    \includegraphics{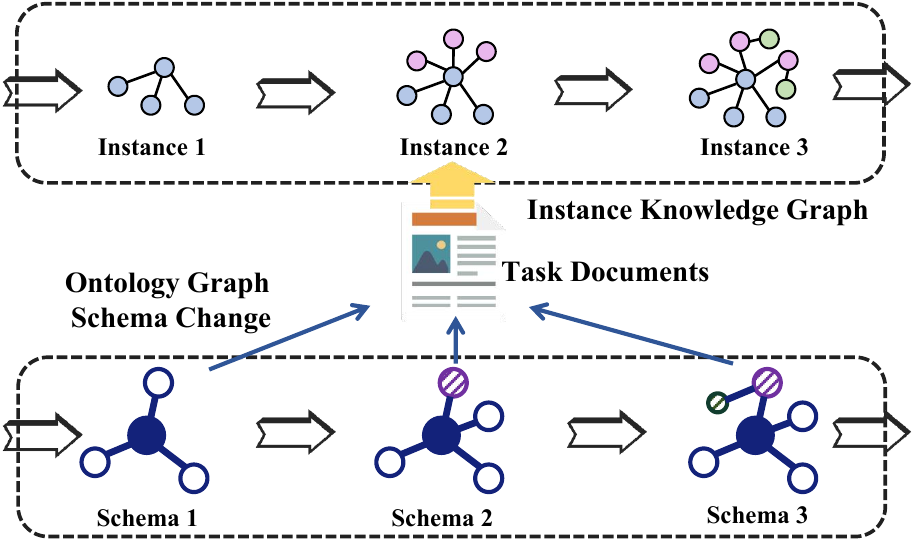}}
    \caption{
    Knowledge Graph Construction (KGC) with dynamic updates of \emph{schema}.}
    \label{fig:intro}
\end{figure}

Knowledge Graph Construction (KGC), typically through information extraction, has enjoyed widespread empirical success and can provide back-end support for
various NLP tasks, such as question answering \citep{DBLP:conf/acl/SaxenaTT20,DBLP:conf/acl/ShangW0022,DBLP:conf/acl/ZhangZY000C22}, commonsense reasoning \cite{DBLP:conf/naacl/YasunagaRBLL21,Zhang2022GreaseLMGR} etc.
Traditional KGC tasks, including named entity recognition (NER) \citep{DBLP:conf/naacl/LiuFTCZHG21,DBLP:conf/acl/WangJBWHHT20a}, relation extraction (RE) \citep{DBLP:conf/sigir/ChenLZTHSC22,DBLP:conf/acl/ZhengWCYZZZQMZ20,DBLP:journals/corr/abs-2210-10709} and event extraction (EE) \citep{DBLP:conf/acl/DaganJVHCR18,DBLP:conf/emnlp/LiuCLBL20,DBLP:conf/acl/0001LXHTL0LC20,DBLP:journals/corr/abs-2301-03282} are ``reactive'',  relying on static pre-defined \emph{schema} from end-users.
However, as shown in Figure \ref{fig:intro}, the schema may evolve along with scenario adaptation, making previous models challenging to utilize without re-training.  

Note that existing information extraction systems can only handle a fixed number of classes by pre-defined schema and performing once-and-for-all training on a fixed framework.
It is desirable to respond to changes (e.g., evolving schema) to existing KGs, making the system act more “proactively” like humans who can handle flexible knowledge updates.
Early, several approaches introduce incremental learning \citep{DBLP:conf/emnlp/CaoCZW20,DBLP:conf/naacl/WangXYGCW19,DBLP:conf/nlpcc/ShenJSCL20,DBLP:conf/acl/CuiYYHCYX20} to learn new classes continually.
In this case, the extraction system learns from the class incremental data stream but usually suffers significant performance degradation on the old class when adapting to the new class. 
Stated differently, previous studies put emphasis on struggling against catastrophic forgetting \citep{DBLP:books/sp/98/Thrun98}.
However, for the schema-evolving scenarios, the dynamic generalizability of extraction models plays a vital role and needs to be inspected from an ontology evolution perspective.

Therefore,  we propose a novel KGC task dubbed schema-adaptable KGC, where the models are required to have the ability to represent and adapt to complement knowledge extraction.
We first construct datasets according to three principles of evolutionary schema directions (\textit{Horizontal Schema Expansion}, \textit{Vertical Schema Expansion}, and \textit{Hybrid Schema Expansion}) on three tasks of NER, RE\footnote{We regard RE as relational triple extraction in this paper.}, and EE.
Through empirical analysis, we notice that approaches of Text2Event \cite{DBLP:conf/acl/0001LXHTL0LC20}, TANL \cite{DBLP:conf/iclr/PaoliniAKMAASXS21}, UIE \cite{DBLP:conf/acl/0001LDXLHSW22}, and GPT-3.5\cite{DBLP:conf/nips/Ouyang0JAWMZASR22} cannot effectively extract the information given complex evolving schema.
We argue that  the major issues lie in the following:
1) How to learn dynamic and generalizable schema representations as conditions for extraction;
2) How to precisely extract new instances constrained with newly updated schema.

\begin{figure*}[t]
    \centering
    \resizebox{.99\textwidth}{!}{
    \includegraphics{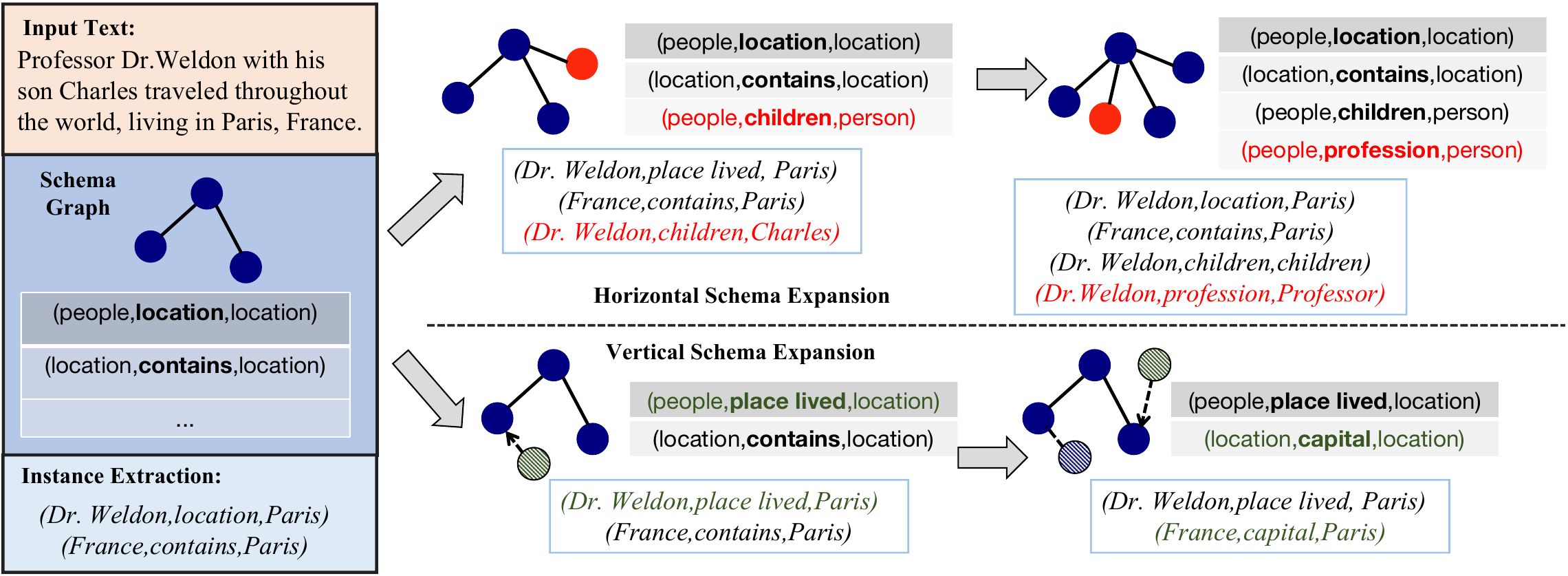}}
    \caption{
    Illustration of schema-adaptable KGC with iterative schema evolution.
    Schema graph nodes (colored \textcolor{DarkBlue}{\CIRCLE}) are evolved by adding a new class node (colored \textcolor{red}{\CIRCLE}) in horizontal schema expansion  while is inherited by a new subclass node
    (colored \textcolor{DarkGreen}{\CIRCLE}) in vertical schema expansion.
    }
    \label{fig:task}
\end{figure*}

To this end, we propose a simple baseline dubbed \textsc{AdaKGC}, which introduces 
\textit{Schema-enriched Prefix Instructors (SPI)} to represent and transfer the learnable schema-specific knowledge. 
At each iteration stage, we linearly convert from the current schema graph to learnable prompts, initialized with the ontology name and connected to task-specific prefixes.
To encourage the decoder to understand the dynamic schema, we utilize a \textit{Schema-conditioned Dynamic Decoding (SDD)} strategy that constructs a decoding path of schema-specific vocabulary to the output space. 
When the schema changes, we dynamically construct a new trie-tree to adjust the output space. 
Note that \textsc{AdaKGC} is model agnostic and can handle a variety of challenging schema evolution scenarios. 
We summarize the contribution of this work as:

\begin{itemize}
\item We introduce a new task of the schema-adaptable KGC to meet the schema evolution requirements,  which is a new branch that has not been well-explored to the best of our knowledge.

\item We propose a new baseline \textsc{AdaKGC}, which includes schema-enriched prefix instructors and schema-conditioned dynamic decoding strategy, and experimentally demonstrate the adaptability. 

\item  We release the schema-adaptable KGC benchmark, which imposes new challenges and presents new research opportunities for the NLP community.
\end{itemize}

\section{Problem Statement and Overview}

\subsection{Background of KGC}

KGC has been a promising research challenge \citep{DBLP:conf/acl/0001LDXLHSW22,DBLP:conf/naacl/ZhangZL022}, and
existing benchmarks utilize a well-defined schema for directing knowledge graph construction, focusing on generating domain-specific knowledge graphs or aggregating heterogeneous structured databases.
For example, FEW-NERD \citep{DBLP:conf/acl/DingXCWHXZL20} consists of coarse-grained and fine-grained entity type definitions to locate and classify named entities from unstructured natural language.
NYT \citep{DBLP:conf/pkdd/RiedelYM10} extracts relational triple instances specifically from textual data sources according to a specific taxonomy structure.
ACE2005 \citep{NtroductionTheA2} identifies triggers and event types based on context, and each has its own event arguments, described in a slot-filling way.
In addition, TAC-KBP \citep{ellis2014overview} is designed to leverage existing generic domain structured data sources and extend entity links employing descriptive text as additional information.
OAEI \citep{DBLP:journals/jods/EuzenatMSSS11} creates an integrated ontology based on an alignment between two or more existing ontologies or knowledge graphs.
In this paper, we focus on the work of extracting knowledge instances from unstructured text, which is regarded as the schema-constraint prediction (structure prediction) task.

\subsection{Definition of Schema-adaptable KGC}

In the real world, the KGC system extracts structured knowledge from unstructured text and normalizes it to the instance graph according to a frequently adjusted schema.  
Given a set of schema graphs $\mathcal{S}=\{s_1, s_2,...,s_n\}$, the task of schema-adaptable KGC is to generate a set of schema-constraint instances $\mathcal{G}=\{g_1, g_2,...,g_n\}$ for each iteration.
Suppose there is a model $\mathcal{M}_{\theta} = LM(\mathcal{D}^{1}_{train}|S_1)$ trained on the initial training set, after which labeled data for updated schema are not available.
A schema-adaptable data stream $\left\{\mathcal{D}^{(1)}, \mathcal{D}^{(2)}, \ldots, \mathcal{D}^{(N)}\right\}$ is provided to evaluate the adaptability of model for the dynamic updates of schema.
Each  $\mathcal{D}_{\text {}}^{(k)}$ contains dev/test data $\left(\mathcal{D}_{\text {dev}}^{k},\mathcal{D}_{\text {test}}^{k}\right)$ and schema graph $s_k$.
Note that the model \textbf{will not be re-trained} but hope to pick up on the ability of information extraction with evolving schema.
The challenge is that the model is expected to perform well in each iteration of the test set $\mathcal{D}_{\text {test}}^{k}$, which contains the golden instances changed for the updated schema.

\subsection{Dataset Construction Process}
\label{sec:dataset-construction-process}
As shown in Figure~\ref{fig:task}, we design three principles regarding different types of schema evolution and apply Algorithm~\ref{alg:main} to build the dataset for evaluation: 
(1) \textbf{Horizontal Schema Expansion} requires the schema to add new class nodes of the same level, which can be considered a form of class-incremental learning without new classe instances as training data. 
Based on the generalization effect on the neighboring new classes, we can assess the transfer capabilities of the schema feature.
(2) \textbf{Vertical Schema Expansion} requires the schema to add subclasses of father classes. 
Based on the generalization effect on subclasses, we can assess the inheritance and derivation capabilities of the schema feature.
(3) \textbf{Hybrid Schema Expansion} requires the schema to randomly expands nodes horizontally or vertically at each iteration, which summarizes schema graphs and represents their potential co-evolutionary pattern. 
More details are in Appendix~\ref{appendix:dataset-construction}, besides the above structural extensions, we further explore analogous node replacement from the perspective of semantics.

\input{algorithm/alg-main.tex}

\subsection{Schema-adaptable KGC Benchmark}
There are two challenges for schema-adaptable KGC. 
Firstly, since the schema is updated in each iteration, the schema evolution information needs to be dynamically injected into the model. Secondly, since the output target of KGC is often demand-specific, the extraction results should be adaptively adjusted according to the schema.
We detail several vanilla baselines as follows and introduce the proposed \textsc{AdaKGC} in \S \ref{sec:adakgc}.

\begin{figure*}[t]
    \centering
    \resizebox{0.99\textwidth}{!}{
    \includegraphics{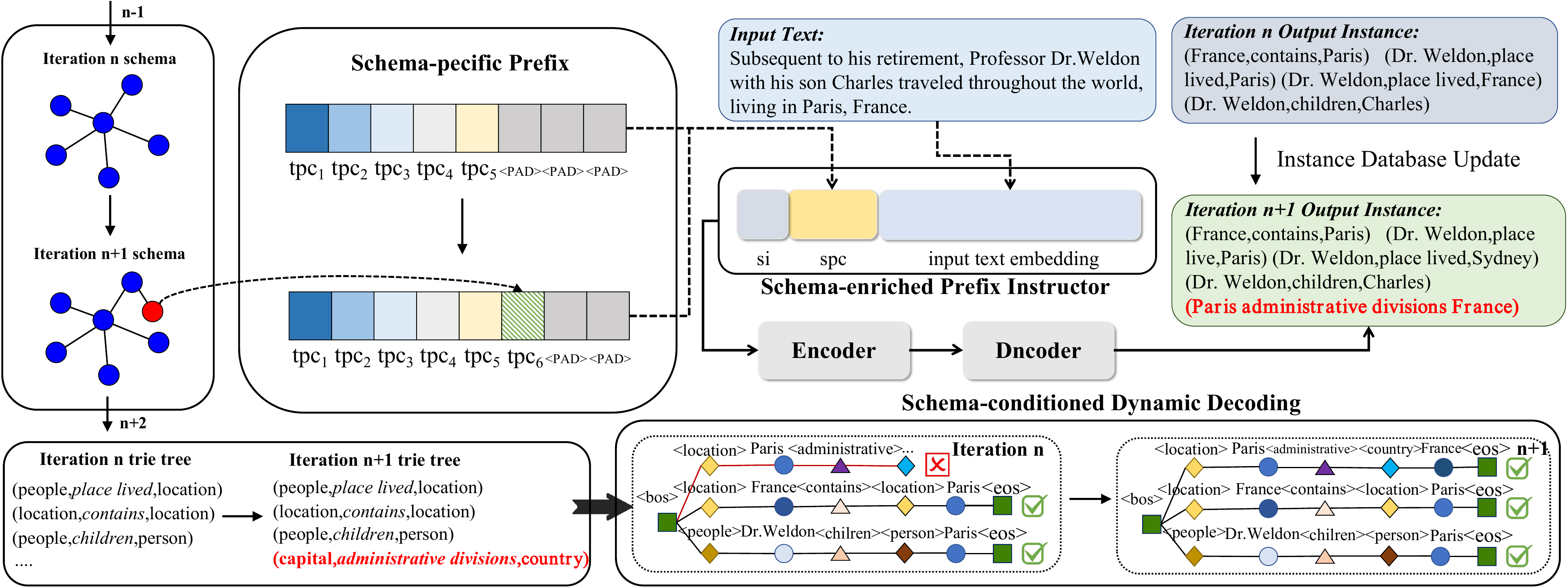}}
    \caption{
    The overview of \textbf{ADA}ptive \textbf{K}nowledge \textbf{G}raph \textbf{C}onstruction \textsc{AdaKGC}.}
\label{fig:model}
\end{figure*}

\paragraph{Vanilla Baselines:}
Schema-adaptable KGC can be thought of as a structured prediction language task that transfers information between class nodes through the generalizability of the structure.
\textbf{TANL} \citep{DBLP:conf/iclr/PaoliniAKMAASXS21}  introduces an augmented natural language translation task from which information related to the schema can be implicitly extracted.
\textbf{TEXT2EVENT} \citep{DBLP:conf/acl/0001LXHTL0LC20} is a unified sequence-to-structure architecture for event extraction with a constrained decoding algorithm for event schema knowledge injection during inference.
\textbf{UIE} \citep{DBLP:conf/acl/0001LDXLHSW22} is a unified text-to-structure generation framework that enables unified modeling of different IE tasks and adaptively generates target sequences by a schema-based prompting mechanism.
\textbf{GPT-3} \cite{DBLP:conf/nips/BrownMRSKDNSSAA20}, a large-scale language model (LLM), can serve as a baseline for schema-adaptable KGC.
Although current works focusing on structured extraction can achieve excellent performance with static types of knowledge, they are typically unaware of schema evolution.
To clarify this issue, we introduce a simple yet effective baseline dubbed schema-\textbf{ADA}ptive \textbf{K}nowledge \textbf{G}raph \textbf{C}onstruction \textsc{AdaKGC}.

\section{The Proposed Baseline: \textsc{AdaKGC}}
 \label{sec:adakgc}

\subsection{Overview}
As shown in Figure~\ref{fig:model}, \textsc{AdaKGC} utilizes a pre-trained encoder-decoder language model (LM) T5 \citep{DBLP:journals/jmlr/RaffelSRLNMZLL20} as the basic architecture for the schema-adaptable KGC task. 
Specifically, let encoder input $X_{en} = [S;X]$ be the concatenation of schema $S$ and input $X$.
In the decoding process, the LM calculates the conditioned probability of generating a new token $y_t$ given the previous token $y_{<t}$: 
\begin{equation}
  p(Y_{de} \mid X_{en})  
  =\prod_{t=1}^{|Y|} p\left(y_t \mid y_{<t}, S,X\right)
\end{equation}

We initialize the model using the pre-trained parameter $\theta$. 
Here, $p_\theta$ is a trainable language model distribution.
In the $k$-th iteration, we perform a gradient update on the following log-likelihood objective:

\begin{equation}
\begin{split}
& \max _\theta \log p_\theta(y \mid x;s_k) \\
& =\max_\theta \sum_{t \in Y_{index}}  \log p_\theta (h_t \mid h_{<t})
\end{split}
\end{equation}

where $h_t$ is the activation vector at decoding time step $t$.
$h_t=\left[h_t^{(1)} ; \cdots ; h_t^{(m)}\right]$ is a concatenation of all activation layers, and $h_t^{(j)}$ is the activation vector of the $j$-th layer at time step $t$.

\subsection{Schema-enriched Prefix Instructor}

Inspired by prefix-tuning \citep{DBLP:conf/acl/LiL20}, we use task-specific prefix instructors to indicate task information,
which are pairs of transformer-activated differentiable sequences $\{si_{en},si_{de}\}$, each containing $p$ consecutive $D$-dim vectors for encoder and decoder.
Since using a discrete natural language task instruction in the context (e.g., "The schema used for the task is:") may guide the LM to produce a sub-optimal generated sequence, we optimize the instructions as a continuous soft prompt, propagating upward to all transformer activation layers and rightward to subsequent tokens.

Due to schema changes with iterations, we present schema-specific prefix instructors to instruct the encoding process.
Specifically, we formalize the schema graph as linearized text. Assume given the constrained schema of RE task $s_k=\{(h_1,r_1,t_1),...(h_n,r_n,t_n)\}$ and $tpc_i=(h_i,r_i,t_i)$ denotes the $i$-th triple prefix constraint. 
By concatenating these schema prefix constraints initialized by word embedding, $spc$ can be dynamically adjusted as the schema evolves, and added padding tokens to be a fixed length when instructing the LM:

\begin{equation}
spc =\text {Concat}\left(tpc_1, \ldots,tpc_n,PAD\right)
\end{equation}

Thus, the schema-enriched prefix instructor provides a two-part prefix combination $Z=\{si_{en}, spc_{c}; si_{de}\}$, where ";" separates the respective prefix instructors for encoder and decoder.
We recursively activate the decoder transformer activation vector $h_t$, which is the connection of all layers, at time step $t$ in the LM.

\begin{equation}
h_t= \begin{cases}  si_t, & \text { if } t\leq
p     \\ \operatorname{LM}\left(y_t, h_{<t} \mid S, X\right), & \text { otherwise }\end{cases}
\end{equation}

The training parameters of our model contain the LM parameters $\theta$, the encoder-decoder task-specific prefix instructor$\{si_{en},si_{de}\}$,
and the schema-specific constraint instructor $spc$.
For stable optimization,  we follow \citet{DBLP:conf/acl/LiL20} to reparameterize the matrix $M_{\phi}[t,:] = MLP_{\phi}(M_{\phi} ' [t,:])$ with a smaller matrix $M_{\phi} '$ consisting of a large feedforward neural network $MLP_{\phi}$, which can alleviate the optimization instability caused by directly updating the prefix parameters and is applied to  $\{si_{en}, si_{de}; spc\}$.
We train the parameters of the model in the following steps:
(1) First, freeze other parameters, fine-tune the prefix instructor
$\{si_{en},si_{de}\}$ to learn task-specific prompts;
(2) Secondly, freeze $\{si_{en},si_{de}\}$, optimize the schema-specific instructor $spc$ for the given schema graph; 
(3) Finally, we unfreeze the LM parameter $\theta$ and collaboratively optimize all parameters to capture the association between the prefix instructor and model parameters.

\subsection{Schema-conditioned Dynamic Decoding}

\input{tables/experiment-horizontal.tex}
\input{tables/experiment-vertical.tex}

\input{tables/experiment-hybrid.tex}

Previous works leverage a greedy decoding algorithm to generate linearized instance predictions token by token for the hidden sequence $h_t$, which selects the token with the highest prediction probability $p\left(y_t \mid y_{<t}, S,X\right)$ at each vanilla decoding step $t$.
Unfortunately, when the schema changes, this decoding algorithm does not guarantee the generated instances are consistent with the latest schema. 
In other words, it may result in out-of-date or invalid types being generated due to the lack of labeled data fine-tuning the model to adapt the probability distribution to the current schema constraints.
In addition, the greedy decoding algorithm neglects useful schema knowledge that can effectively guide the decoding process. 

In the schema-conditioned decoding process, we apply a trie-based decoding mechanism that dynamically constructs a trie-tree by leveraging the latest schema. 
An intuitive interpretation is that the schema contains rich semantic information  (\eg, instance types) and structural information (\eg, relational edges between instance types) so that the decoding process can be constrained to ensure that the generated token is valid.
Specifically,  we constrain the model to generate the type tokens consistently with the existing schema at the type location.
We pursue the LM output to be a sequence of RE following pattern and optimized using the standard seq2seq objective function:
\begin{center}
$\begin{array}{c}
[bos] \ldots
\mathcal{T}_{h}^{(n)},\mathcal{E}_{h}^{(n)}, \mathcal{R}^{(n)}, \mathcal{T}_{t}^{(n)}, \mathcal{E}_{t}^{(n)}\ldots [eos] \\
\end{array}$  
\end{center}

where $\mathcal{E}_{h}^{(n)}$, $\mathcal{T}_{h}^{(n)}$, $\mathcal{E}_{t}^{(n)}$, $\mathcal{E}_{h}^{(n)}$ refer to the $n$-th generated head entity, tail entity, and their respective types while $\mathcal{R}^{(n)}$ refer to relation.

\renewcommand{\algorithmicrequire}{\textbf{Input:}}
\renewcommand{\algorithmicensure}{\textbf{Output:}}

\section{Experiments}
\label{Experiments}


\subsection{Experimental Settings}

\noindent
\textbf{Datasets.}
We conduct experiments on KGC tasks, including NER, RE and EE.
The used datasets includes FEW-NERD \citep{DBLP:conf/acl/DingXCWHXZL20}, NYT \citep{10.1007/978-3-642-15939-8_10} and
ACE2005 \citep{ace2005-annotation}.
In our work, we need to construct schema as well as golden validation/test sets dynamically. 
For each dataset, we build three types of evaluation settings based on \S \ref{sec:dataset-construction-process}.
Therefore for original datasets, we use a certain proportion of the schema as initialization to conduct schema expansion regarding three schema evolution categories in Appendix~\ref{appendix:dataset-construction}.

\noindent
\textbf{Evaluation.}
We use span-based Micro-F1 as the primary metric.
\textbf{Rel-S} means that the relation is correct if the relation type is correct and the string and entity types of the related entity mentions are correct.
For each iteration experiment, we report the average performance over 3 random seeds.
UIE is implemented without pre-training by directly using T5-v1.1-base as the backbone for a fair comparison.
More details are in Appendix~\ref{appendix:evaluation}.

\subsection{Main Results}

We report empirical results regarding horizontal schema expansion, vertical schema expansion and hybrid schema expansion settings to compare our proposed methods with the baselines.
The performance over all iterations during the whole schema-adaptable KGC process is presented in Table~\ref{tab:Horizontal-experiment}-\ref{tab:hybrid-experiment}.  
From the results, we can observe that:

\noindent
\textbf{Schema adaptive generalization challenge.}
On all three expansion categories, the model performances tend to decrease as the iterations increase.
\textsc{TANL} achieves lower performance which employs an augmented language and implicitly trains the model to learn schema information.
\textsc{TEXT2EVENT} utilizes schema as constraint information on the decoding side and outperforms other models in some iterations.
Although \textsc{AdaKGC} and \textsc{UIE} obtain optimal or suboptimal performance, the performance of iteration 1 and iteration 7 has a significant drop.
We believe that the implicit schema evolution rules can help future work to develop adaptive generalization capabilities for schema-adaptable KGC.

\begin{figure}[t]
    \centering
    \resizebox{.49\textwidth}{!}{
    \includegraphics{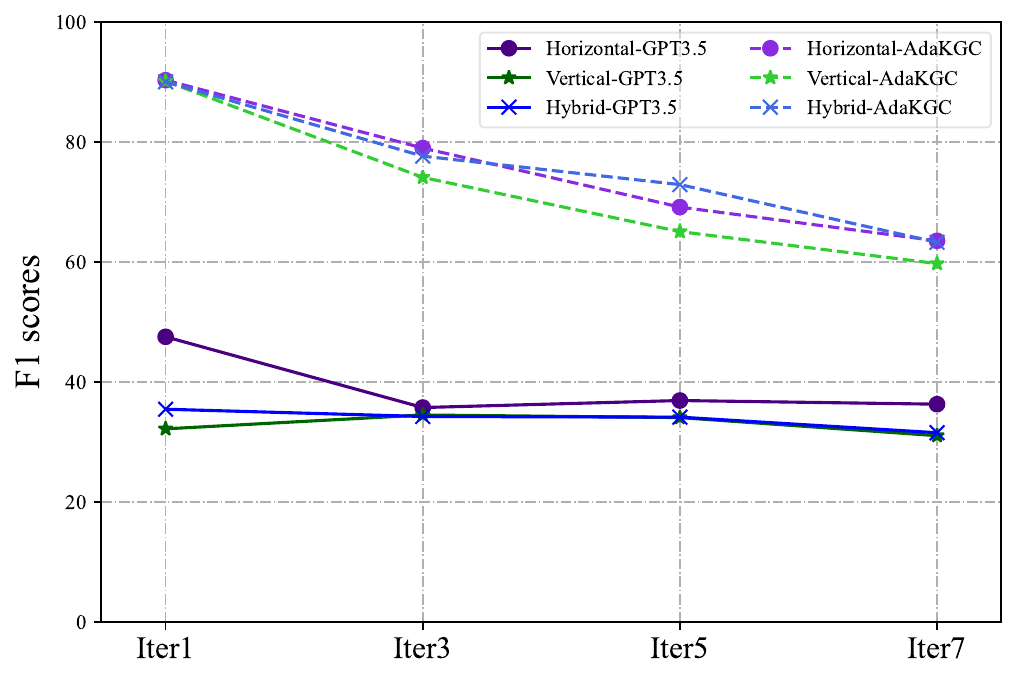}}
    \caption{
    GPT-3.5 results on schema expansion dataset. }
    \label{fig:case-gpt3.5}
\end{figure}

\begin{figure}[t]
    \centering
    \resizebox{.49\textwidth}{!}{
    \includegraphics{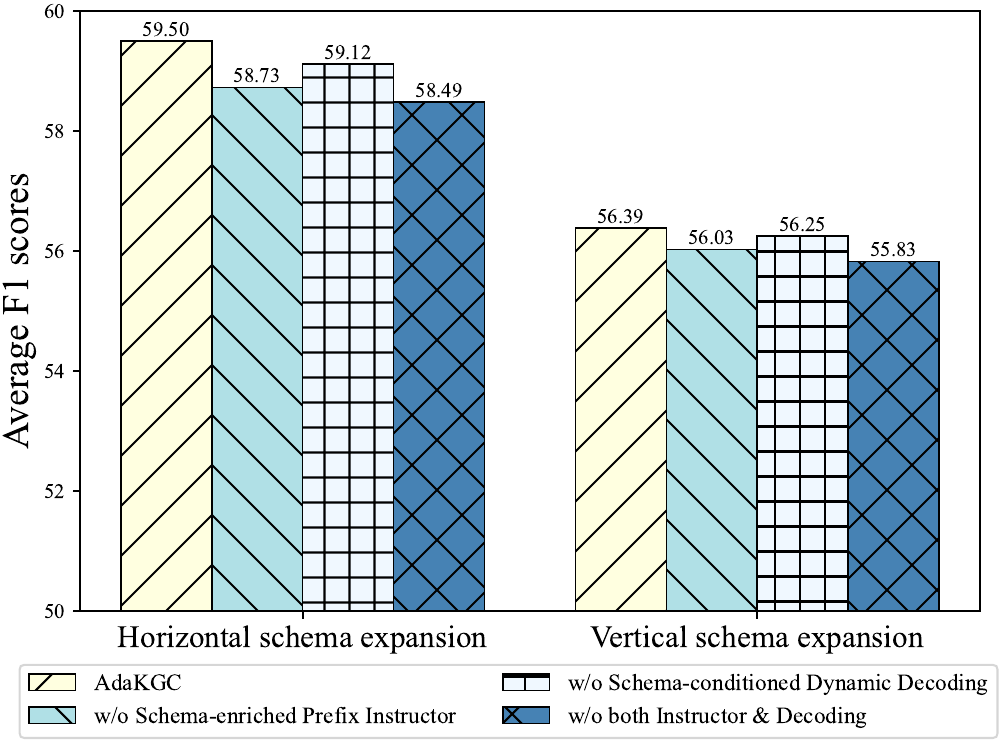}}
    \caption{
    Ablation study on NERD horizontal schema expansion dataset, with the average result of 7 iterations. }
    \label{fig:ablation}
\end{figure}

\noindent
\textbf{Schema-enhanced modules boost the performances.}
Compared to other models, \textsc{AdaKGC} is improved with schema-enhanced modules on both the encoder and decoder, which allows it to achieve the best performance in most settings.
On the  ACE2005 hybrid schema expansion dataset, \textsc{AdaKGC} improves 0.71\% on trigger extraction and 3.65\% on event argument extraction, indicating that \textsc{AdaKGC} can capture schema-specific information under evolutionary schema.

\noindent
\textbf{LLMs can understand schema adaption patterns better.} 
To explore the performance of LLMs \citep{DBLP:journals/corr/abs-2212-09597}  on the proposed tasks, we perform comparative experiments with \textsc{GPT-3.5} on NYT. 
Since we cannot utilize all training instances, we report in-context learning performance given 20-shot demonstrations as shown in Appendix~\ref{appendix:gpt3.5}.
From Figure~\ref{fig:case-gpt3.5}, we notice that \textsc{GPT-3.5} is capable of producing instances that conform to the dynamically changing schema but still yield low performance due to the low-shot issue.
Likewise, we sample several cases and use ChatGPT\footnote{\url{https://openai.com/blog/chatgpt/}} to evaluate schema-adaptable KGC (See Figure \ref{fig:chatgpt1} and \ref{fig:chatgpt2} in Appendix \ref{appendix:chatgpt}), which surprisingly demonstrates stable generalization ability with evolving schema.
These findings indicate a promising future work of schema-adaptable KGC to develop alignment prompts with LLMs.

\begin{figure}[t]
    \centering
    \resizebox{.49\textwidth}{!}{
    \includegraphics{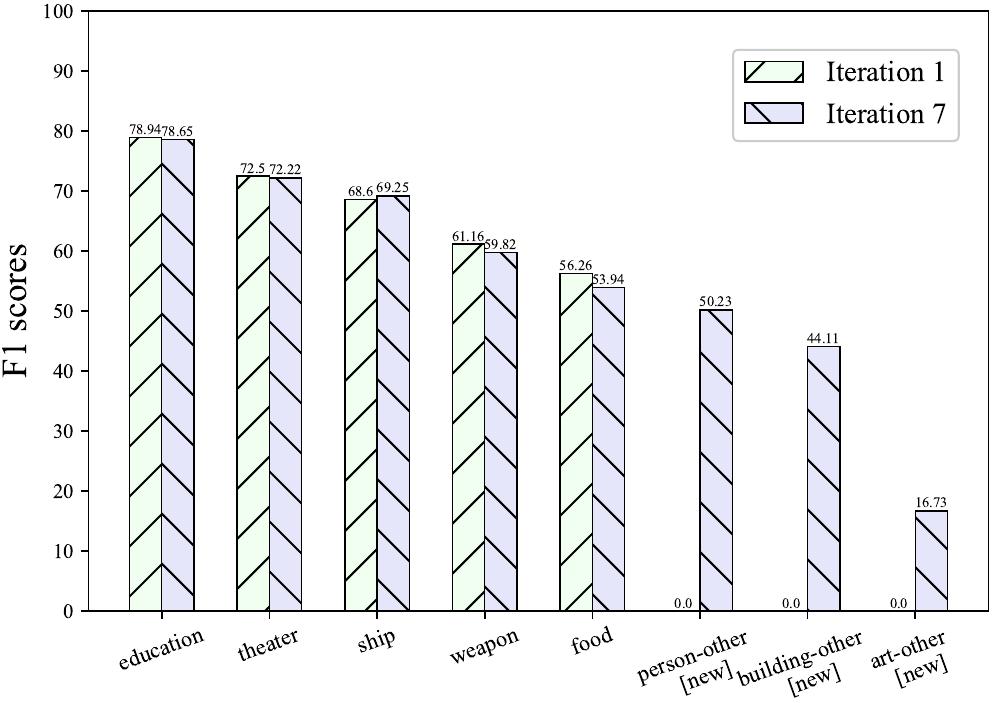}}
    \caption{
    Case study on NERD vertical schema expansion dataset, Iteration 1 vs. Iteration 7. }
    \label{fig:case-study}
\end{figure}

\subsection{Ablation Study on \textsc{AdaKGC}}

To prove the effects of the schema-enriched prefix instructor and schema-conditioned dynamic decoding, we conduct the ablation study, and the results are shown in Figure~\ref{fig:ablation}.
From two evolutionary categories, we observe that: 
(1) Both schema-enriched prefix instructor and schema-conditioned dynamic decoding can help the schema-adaptable learning process;
(2) Efficiently encoding schema evolution information is more important, which achieves improvements of 0.77\% on horizontal schema expansion and 0.36\% on vertical schema expansion.

\input{tables/errorcase-table.tex}

\subsection{Case Study}

As shown in Figure~\ref{fig:case-study}, we randomly select 8 types and observe that:
(1) The types that appear in the initial schema mostly degrade performance, indicating that the model causes slight confusion as the schema expands.
(2) Due to the structural inheritance relationship in the vertical expansion of the schema, our model can effectively transfer the labels of the father node to the child nodes when new child nodes are added.
 
To further analyze the drawbacks of our model and promote future works of schema-adaptable KGC, we count incorrect instances and classify them into five categories below, 
as shown in Table \ref{tab:error-analysis}:
(1) \textit{Weak Transfer.} 
Despite schema expansion, the model is prevented from updating labels by old model parameters.
(2) \textit{Inheritance Deficiency.} 
The label is not inherited in time when subdividing the father node.
(3) \textit{Relevance Neglect.} 
The lack of ontology relevance leads to the absence of correlated event extraction.
(4) \textit{Class Imbalance.} 
Models suffering from unbalanced class learning problems tend to depend on similarly in-context sentences to judge high-frequency labels.
(5) \textit{Potential Annotation.} 
Some example outputs suggest potential errors or omitted annotation.

\section{Related Work}

\subsection{Knowledge Graph Construction}

Automatic construction of knowledge graphs from textual or structured data has attracted extensive research in recent years, including tasks such as NER \citep{DBLP:conf/iclr/PaoliniAKMAASXS21,DBLP:conf/acl/CuiWLYZ21}, RE \citep{DBLP:conf/acl/LinJHW20,DBLP:journals/tacl/JoshiCLWZL20,DBLP:conf/aaai/YeZDCTHC21}, EE \citep{DBLP:conf/emnlp/RamponiGLP20,DBLP:conf/emnlp/LiuCLBL20,DBLP:conf/acl/0001LXHTL0LC20}, etc.
In contrast to closed-domain knowledge extraction, open knowledge extraction \citep{DBLP:conf/acl/KolluruARMC20,DBLP:conf/aaai/ZhanZ20,DBLP:conf/acl/KotnisGRSRTNL22,DBLP:conf/acl/WangPSL22} is oriented toward the absence of schema constraints and can quickly generate extensive and meaningful knowledge.
However, the ignoring of schema introduces uncertainty and ambiguity in output control, and we believe that a clear setting can be chosen to track the realignment of instances. 
Besides, KGC in low-resource scenarios \citep{DBLP:conf/acl/DaganJVHCR18,DBLP:conf/www/ZhangDSCZC20,DBLP:conf/eacl/SchickS21,DBLP:conf/www/ChenZXDYTHSC22,DBLP:conf/www/YeZDCCXCC22,zny2022knowledge} requires the model to predict new instances with only limited training instances available.
As opposed to this instance-driven KGC approach, we argue that the schema-driven approach can leverage evolutionary instructions provided with richer ontological associations, resulting in new challenges and research opportunities.

\subsection{Lifelong Learning}

Lifelong learning is aimed at training new classes online without catastrophic forgetting.
Generally, lifelong learning mainly falls into four categories:
regularization-based \citep{kirkpatrick2017overcoming,DBLP:conf/icml/ZenkePG17,DBLP:conf/eccv/AljundiBERT18},
replay-based \citep{DBLP:conf/nips/Lopez-PazR17,DBLP:conf/nips/ShinLKK17},
architecture-based \citep{DBLP:conf/cvpr/MallyaL18,DBLP:conf/iclr/YoonYLH18}
and knowledge distillation \citep{DBLP:conf/emnlp/ChuangSC20,DBLP:conf/emnlp/CaoCZW20}.
To study class-incremental learning, \citet{DBLP:conf/aaai/MonaikulCFR21} builds a unified NER classifier for all the classes encountered over time, while \citet{DBLP:conf/acl/Wang0ZKMZH22} develops a framework to reconstruct synthetic training data of the old classes.
Recently, \cite{DBLP:conf/naacl/WangXYGCW19} proposes a lifelong RE method that employs an explicit alignment model to overcome forgetting, while \cite{DBLP:conf/nlpcc/ShenJSCL20} presents a self-adaptive dynamic regularization method.
To address class incremental learning in event detection \citep{DBLP:conf/emnlp/CaoCZW20},
\citet{DBLP:conf/emnlp/YuJN21} takes advantage of rich correlations among ontology types, and 
\citet{DBLP:journals/corr/abs-2204-07275} adopts continuous prompts to learn event-specific representation for prediction.
Compared with previous work that focused only on class increments, we discuss three principles of schema expansion from the potential demand for schema adaptation.

\section{Conclusion}
This paper introduces a new task of schema-adaptable KGC with benchmark datasets and a new baseline \textsc{AdaKGC}. 
We illustrate the task difficulties with previous baselines on three principles of schema expansion patterns (horizontal, vertical, hybrid) and demonstrate the effectiveness of the proposed \textsc{AdaKGC}. 

\section*{Acknowledgment}
We would like to express gratitude to the anonymous reviewers for their kind comments. 
This work was supported by the National Natural Science Foundation of China (No.62206246), Zhejiang Provincial Natural Science Foundation of China (No. LGG22F030011), Ningbo Natural Science Foundation (2021J190), Yongjiang Talent Introduction Programme (2021A-156-G), CCF-Baidu Open Fund, and Information Technology Center and State Key Lab of CAD\&CG, Zhejiang University.

\section*{Limitations}
\label{adx:limitation}

The proposed work still contains several limitations, as follows:

\paragraph{Datasets:}

Note that several datasets, such as ACE2005, cannot be released due to LICENCE issues; we release the code to build the datasets and provide all the pre-processed publicly available datasets (e.g., Few-NERD, NYT)
We use several existing datasets to construct schema-adaptable benchmarks; however, previous datasets may have limited schema structures (the schema pattern in some datasets is very simple).
We plan to build more datasets via crowdsourcing for comprehensive evaluation. 
In addition, we will continue to promote the construction of multimodal schema adaptive graphs, which leverage the dynamic evolution of schema to integrate visual and textual knowledge into a self-learning graph extraction system.

\paragraph{Baselines and Proposed \textsc{AdaKGC}:}

Note that the proposed one, although better than previous approaches, including Text2Event \cite{DBLP:conf/acl/0001LXHTL0LC20}, TANL \cite{DBLP:conf/iclr/PaoliniAKMAASXS21}, UIE \cite{DBLP:conf/acl/0001LDXLHSW22}, still suffers from poor generalization ability.
However, we notice a very stable performance with LLM (though deficient performance), indicating a new promising solution for schema-adatable KGC.

\section*{Ethical Considerations}

\paragraph{Intended use.}

The dataset and model in this paper are indented to be used for exploratory analysis of schema-adaptable KGC.

\paragraph{Biases.}
We collect data from existing datasets (e.g., Few-NERD: CC BY-SA 4.0 license.), which may contain some data with offensive language or discriminatory.

\bibliography{anthology,custom}
\bibliographystyle{acl_natbib}

\appendix

\section{Appendix}
This section describes the details of experiments, including dataset construction and evaluation on downstream tasks.

\subsection{Dataset Construction}
\label{appendix:dataset-construction}
\subsubsection{Construction Process}
In each task, we execute three schema evolution strategies. 
The raw dataset statistics are shown in Table ~\ref{tab:details_datasets}, where it can be seen that they have a two-level schema structure, leaving the research of a more hierarchical schema structure for future work.
As shown in Algorithm~\ref{alg:horizontal}-\ref{alg:hybrid},
we describe in detail the specific construction process of horizontal schema extension, vertical schema extension and hybrid schema extension.

In particular, we also release additional datasets from a semantic substitution perspective.
As shown in Algorithm~\ref{alg:analougous}, analogous schema expansion requires schema replacement for semantically similar new nodes. Based on the performance of the old class transfer to the new semantic class, we can evaluate the semantic invariance capability.

\input{tables/raw_datesets_statistics.tex}

\noindent
\textbf{Horizontal Schema Expansion.}
Neighboring nodes of the specified type that have high-level similarity values in the same framework are also adjacent when projected into the semantic space \citep{DBLP:conf/acl/DaganJVHCR18}. 
Existing research efforts have developed many rich libraries of ontologies (\eg FrameNet \citep{DBLP:conf/acl/BakerS03}, VerbNet \citep{DBLP:journals/lre/KipperKRP08}, Propbank \citep{DBLP:journals/coling/PalmerKG05}, and OntoNotes \citep{DBLP:conf/semco/PradhanHMPRW07}),
where each ontology type is associated with a set of pre-defined neighboring ontologies. 
(1) Searching the ontology library to retrieve candidate nodes $W_f$ associated with target nodes $W_s$ at the same hierarchy.
(2) The similarity metric is obtained by calculating the cosine vector similarity of all candidate nodes $W_f$ to the target node $W_s$ (Eq.~\ref{similar1}). (3) Selecting the appropriate threshold of node pairs to confirm the sorted addition of horizontal nodes.
(4) Updating the schema with horizontal nodes and adding the golden validation set and test set in the dataset.

\begin{equation}
\phi_{Sim}(\vec{W_f}, \vec{W_s})=\frac{\sum_{i=1}^{|W_s|} w_i \cdot w_f}{\|\vec{W_f}\|_2 \cdot\|\vec{W_s}\|_2}
\label{similar1}
\end{equation}

\noindent
\textbf{Vertical Schema Expansion.}
Structural similarity needs to be exploited when adding schema hierarchy nodes as new classes. 
(1) For search convenience, we link the hypernym ontology under a root node so that the schema forms a tree structure.
(2) Starting at the root node, we utilize a child selection strategy by recursively applying through the tree until reaching the deepest node. 
A node could be expandable when it represents a non-terminal state or has hyponyms in semantics (e.g., location->country).
(3) According to the available hyponyms, one (or more) child nodes are added to expand the current schema tree.
(4) Updating the schema with vertical nodes and adding the golden validation set and test set in the dataset.

\noindent
\textbf{Hybrid Schema Expansion.}
It is necessary to hybrid horizontal and vertical expansion to form a comprehensive structural topology, which is more consistent with real scenarios.
(1) Setting
the threshold $\alpha$ for random selection. 
(2) Executing horizontal expansion iteration below the threshold  $\alpha$, or  vertical node expansion above the threshold  $\alpha$.
Note that when the father node of added nodes does not exist, we also add the father node to maintain the schema hierarchy.
(3) Updating the schema with the corresponding nodes and adding the golden validation set and test set in the dataset.

\noindent
\textbf{Analogous Schema Expansion.}
To detect the semantic node sensitivity of the schema, we randomly replace similar semantic expressions for the nodes.
(1) Random selection of candidate nodes to obtain word expressions $W_C$. 
(2) Candidate nodes are created by pairing $W_C$ with all words in the corpus word list $W_L$. The consistency between individual words is calculated by the normalized point-by-point mutual information (NPMI) of $w_i$ and $w_j$ (Eq. ~\ref{nmpi1}), where adding smooth $\epsilon$ and $\gamma$ controls for $\log p\left(w_i, w_j\right)$ weights for higher NPMI values (Eq. ~\ref{nmpi2}).
(3) Adopting candidate nodes that exceed the threshold to replace the schema and updating the golden validation set and test set in the dataset.

\begin{equation}
\vec{v}(W)=\left\{\sum_{w_i \in W_C} NMPI\left(w_i, w_j\right)^\gamma\right\}_{j=1, \ldots,|W_L|}
\label{nmpi1}
\end{equation}

\begin{equation}
N M P I\left(w_i, w_j\right)^\gamma=\left(\frac{\log \frac{p\left(w_i, w_j\right)+\epsilon}{p\left(w_i\right) \cdot p\left(w_j\right)}}{-\log p\left(w_i, w_j\right)+\epsilon}\right)^\gamma
\label{nmpi2}
\end{equation}

\subsubsection{Schema-adaptable Datasets Statistic}
We set the number $\mathcal{N}$ of total iterations, and initialize the original number of schema nodes. We show the statistics of schema-adaptable datasets for each task in Table~\ref{tab:Horizontal node expansion datasets}.

\input{tables/schema-adaptable-statistics.tex}

\subsection{Evaluation}
\label{appendix:evaluation}
We use span-based Micro-F1 as the major metric to evaluate the model and  adopt the same evaluation metrics as previous work:
\begin{itemize}
\item[*] \textbf{Named Entity Recognition}: an entity mention is correct if its strings and type match a reference entity.
\item[*] \textbf{Relation Strict}: a relation is correct if its relation type is correct and the string and entity types of the related entity mentions are correct.
\item[*] \textbf{Event Trigger}: an event trigger is correct if its strings and event type match a reference trigger.
\item[*] \textbf{Event Argument}: an event argument is correct if its strings, role type, and event type match a reference argument mentioned.
\end{itemize}

\subsection{Hyper-parameters}

We adopt T5-v1.1-base \citep{2020t5}, which has 12 layers of the encoder, 12 layers of the decoder, 768 hidden units, and 12 attention heads as the backbone. 
Specifically, we utilize Pytorch \cite{paszke2017automatic} to conduct experiments with batch size 16 on one NVIDIA 3090 GPU.
We detail the hyper-parameters for each dataset as follows: 

\textbf{NERD.}
The hyper-parameter search space is:
\begin{itemize}
\item epoch: 15
\item batch size: 16
\item accumulate: 1
\item learning rate: [\textbf{1e-4}, 3e-4, 5e-4]
\item warmup rate: 0.06
\end{itemize}

\textbf{NYT.}
 The hyper-parameter search space is:
\begin{itemize}
\item epoch: 20
\item batch size: 16
\item accumulate: 1
\item learning rate: [\textbf{1e-4}, 3e-4, 5e-4]
\item warmup rate: 0.06
\end{itemize}

\textbf{ACE2005.}
 The hyper-parameter search space is:
\begin{itemize}
\item epoch: 30
\item batch size: 16
\item accumulate: 1
\item learning rate: [\textbf{1e-4}, 3e-4, 5e-4]
\item warmup rate: 0.06
\end{itemize}

\subsection{Analogous Schema Expansion Experiment}

As shown in Figure~\ref{tab:Analogous-experiment}, our \textsc{AdaKGC}
also has powerful semantic transplantation capabilities,
which achieves competitive performance with baselines.
With the schema-enriched prefix instructor, \textsc{AdaKGC} achieves an improvement of 7.70\% on average over TEXT2EVENT on the event trigger extraction task and 4.87\% on the event argument extraction task. 
This verifies the proposed schema-enriched prefix instructor and decoding modules can learn general schema-adaptable ability even the schema evolution knowledge is rarely in the pre-training stage.
Note that \textsc{TANL} achieves the best performance on the NYT dataset, indicating that language models have the ability to learn schema semantic transfer implicitly as an augmented natural language prediction task.
Therefore we believe that in addition to the schema structure perception modules, semantic robustness modules for analogous node expansion scenarios are also essential.

\input{algorithm/alg-horizontal.tex}

\input{algorithm/alg-vertical.tex}

\input{algorithm/alg-hybrid.tex}

\input{algorithm/alg-analogous.tex}

\subsection{GPT-3.5 Experiment Details}
\label{appendix:gpt3.5}

GPT-3.5 is a large autoregressive language model with 175 billion parameters.
To explore the performance of GPT-3.5 on the schema-adaptive KGC task, we follow the input format of few-shot learning using OpenAI API\footnote{\url{https://platform.openai.com/docs/models/gpt-3-5}}.
As shown in Table~\ref{tab:gpt3.5-input},
we utilize a fixed manual template to generate a contextual window suitable for the model, including natural language task descriptions (text in \textcolor{blue}{blue}), linearized schemas (text in \textcolor{purple}{purple}),  20 examples in the model’s context, and task prompts (text in \textcolor{red}{red}).

\subsection{ChatGPT Results}
\label{appendix:chatgpt}

\begin{figure*}[t]
    \centering
    \resizebox{.98\textwidth}{!}{
    \includegraphics{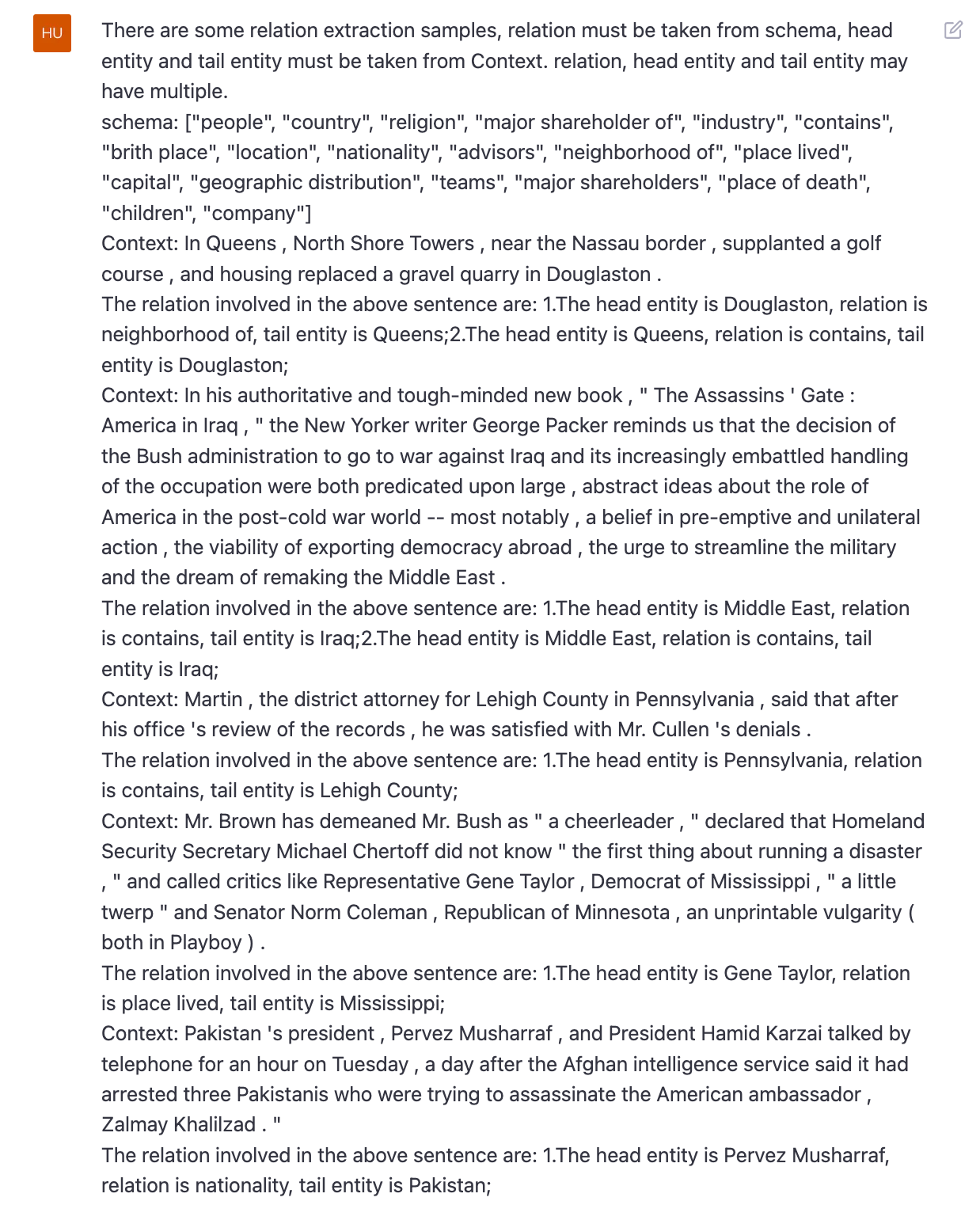}}
    \caption{
    ChatGPT input example on NYT dataset.  A total of 20 demonstrations are given to the model.}
    \label{fig:chatgpt1}
\end{figure*}

\begin{figure*}[t]
    \centering
    \resizebox{.98\textwidth}{!}{
    \includegraphics{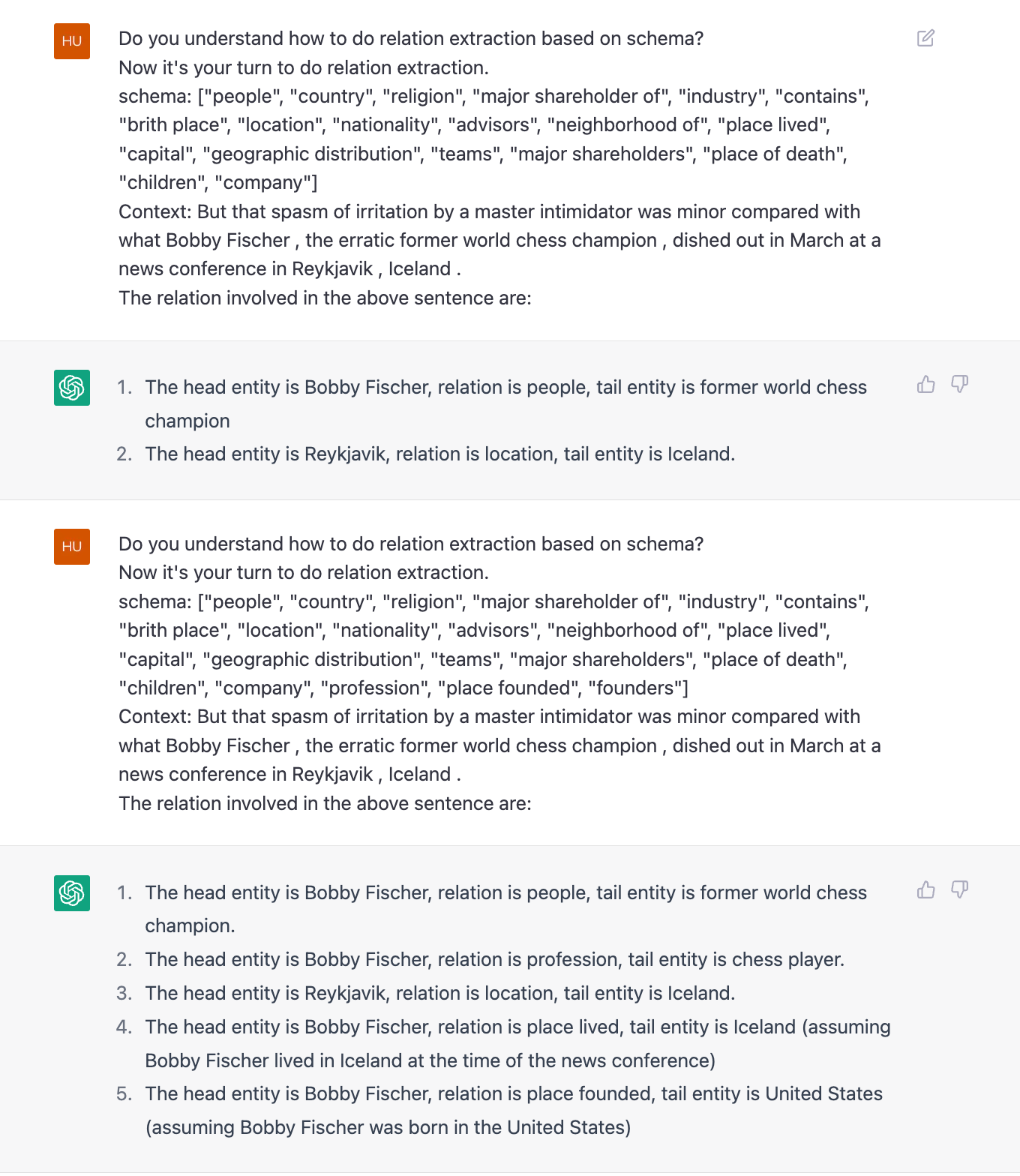}}
    \caption{
    ChatGPT output results under a horizontal schema expansion iteration. }
    \label{fig:chatgpt2}
\end{figure*}

ChatGPT\footnote{\url{https://chat.openai.com/chat}} trains an initial model using supervised fine-tuning and further utilizes reinforcement learning systems to rank by quality for human feedback rewards. 
We handle Schema-adaptable KGC tasks by asking questions to the chatbot in a conversational mode.
First, we present the task description and the 20 demonstrations as shown in Figure~\ref{fig:chatgpt1}.
Then we give a paragraph text to test whether the chatbot can extract the corresponding triples based on the same schema as the demonstration examples comply with.
From Figure~\ref{fig:chatgpt2} we can find that some of the facts are well extracted, indicating that ChatGPT can understand the task and perform extraction consistent with the schema.
Finally, we add three new nodes \textit{“profession” “place founded” "founders"} to the previous schema under a horizontal schema expansion iteration.
Output results in Figure~\ref{fig:chatgpt2} show that ChatGPT not only adapts the output to the updated schema but also deduces reasonable facts by a chain-of-thought approach.

\input{tables/experiment-analogous.tex}

\input{tables/data-example-hor-ace.tex}

\input{tables/data-example-ver-ace.tex}

\input{tables/data-example-mix-ace.tex}

\input{tables/data-example-ana-ace.tex}

\input{tables/gpt3-5.tex}

\end{document}

%% file: algorithm/alg-main.tex
\begin{algorithm}[t]
\caption{Dataset Construction Process.}  
\label{alg:main} 
{\bf Input:}   iteration $\mathcal{N}$, raw schema $\mathcal{S}_{raw}$, and raw dataset $\{\mathcal{D}_{train}^{raw},\mathcal{D}_{dev}^{raw},\mathcal{D}_{test}^{raw}\}$

{\bf Output:}  Schema $\mathcal{S}_\mathcal{N}$,$\{\mathcal{D}_{train}^\mathcal{N},\mathcal{D}_{dev}^\mathcal{N},\mathcal{D}_{test}^\mathcal{N}\}$

\begin{algorithmic}[1]
\STATE Randomly initialize $n_{init}$ nodes in
$\mathcal{S}_{raw}$ as $\mathcal{S}_1$

\STATE Pick out the instance associated with schema $\mathcal{S}$ as $\mathcal{D}^{(1)} =\{ \mathcal{D}_{train}^1,\mathcal{D}_{dev}^1,\mathcal{D}_{test}^1\}$

\FOR{iteration $i=2,\dots,n$}    

\STATE  \textbf{Horizontal Schema Expansion:} Compute 
$\phi_{neighbor}(\vec{W_c}, \vec{W_s})$ for candidate schema $\mathcal{S}$ 

\STATE  \textbf{Vertical Schma Expansion:} 
Select $n_{iter}$ sub node whose \emph{father node} belongs to $\mathcal{S}$ and update

\STATE  \textbf{Hybrid Schema Expansion:} 
Combine extension Steps 4 and 5

\STATE Ouput iteration $i$ dataset schema $\mathcal{S}_i = \mathcal{S}$, instance  $\mathcal{D}^{(i)} = \{\mathcal{D}_{dev}^i,\mathcal{D}_{test}^i\}$

\ENDFOR

\end{algorithmic}
\end{algorithm}

%% file: tables/experiment-horizontal.tex
\begin{table*}[tbph]
\centering
  \resizebox{0.95\textwidth}{!}{
\begin{tabular}{cl|cccc ccc|c}
\toprule
      & \multicolumn{1}{c|}{\textbf{Model}} & \textbf{Iter 1} & \textbf{Iter 2} & \textbf{Iter 3} & \textbf{Iter 4} & \textbf{Iter 5} & \textbf{Iter 6} & \textbf{Iter 7} & \textbf{AVE} \\
\hline 
\multicolumn{1}{c}{\multirow{3}[2]{*}{\shortstack{\textbf{Entity} \\ (\textbf{NERD}) \\ \textbf{Ent-F1}}}}
      & TANL
      &  71.52  & 65.21   & 60.38  & 56.37   & 53.21 & 49.66   & 46.55   & 57.56    \\
      & UIE 
      &  72.72  & 66.78   & 62.24  & 58.29   & 55.08 & 51.42   & 48.04   &  \underline{59.22}   \\
      & AdaKGC 
      &  72.91  & 66.95   & 62.37  & 58.51   & 55.38 & 51.81   & 48.54   & \textbf{59.50}   \\
\hline
\multicolumn{1}{c}{\multirow{3}[2]{*}{\shortstack{\textbf{Relation} \\ (\textbf{NYT}) \\ \textbf{Rel-S F1}}}} 
      & TANL 
      & 89.92  & 81.81   & 78.34   & 72.71  & 68.42 & 65.19  &  62.94  &  74.19  \\
      & UIE 
      & 90.17  & 82.09   & 78.74   & 73.12  & 69.03 & 65.53  &  63.30   & \underline{74.57}   \\
      & AdaKGC 
      & 90.34  & 82.33   & 79.03   & 73.34  & 69.19 & 65.87  &  63.58   & \textbf{74.81}    \\
\hline
\multicolumn{1}{c}{\multirow{3}[2]{*}{\shortstack{\textbf{Event Trigger} \\ (\textbf{ACE2005}) \\
\textbf{Evt Tri F1}}}} 
      & TEXT2EVENT 
      & 69.23  & 68.05  & 65.45  & 61.37  & 60.15  & 59.34  & 54.42   &  62.57  \\
      & UIE 
      & 70.75  & 69.13  & 66.20  & 62.19  & 60.90  & 59.83  & 54.74   & \underline{63.39}    \\
      & AdaKGC 
      & 72.43  & 70.90  & 68.14  & 63.49  & 61.97  & 61.33  & 55.73   & \textbf{64.86}    \\
\hline
\multicolumn{1}{c}{\multirow{3}[2]{*}{\shortstack{\textbf{Event Argument} \\ (\textbf{ACE2005}) \\ \textbf{Evt Arg F1}}}}
      & TEXT2EVENT 
      & 46.15  & 44.40  & 42.58   & 39.73  & 39.17 & 38.77  & 35.31  &  40.87  \\
      & UIE 
      & 49.14  & 47.90  & 45.75   & 42.32  & 41.83  & 41.27  & 37.60  & \underline{43.69} \\
      & AdaKGC 
      & 49.18  & 48.14  & 47.08   & 43.85  & 43.17  & 43.10  & 38.79  & \textbf{44.76}   \\

\bottomrule
\end{tabular}%

}
\caption{
    Horizontal schema expansion results in schema-adaptable knowledge graph construction. 
}
\label{tab:Horizontal-experiment}

\end{table*}

%% file: tables/experiment-vertical.tex
\begin{table*}[ht]
\centering
  \resizebox{0.95\textwidth}{!}{
\begin{tabular}{cl|cccc ccc|c}
\toprule
      & \multicolumn{1}{c|}{\textbf{Model}} & \textbf{Iter 1} & \textbf{Iter 2} & \textbf{Iter 3} & \textbf{Iter 4} & \textbf{Iter 5} & \textbf{Iter 6} & \textbf{Iter 7} & \textbf{AVE} \\
\hline
\multicolumn{1}{c}{\multirow{3}[2]{*}{\shortstack{\textbf{Entity} \\ (\textbf{NERD}) \\ \textbf{Ent-F1}}}}
      & TANL
      & 73.49    &  64.83   & 57.05   & 51.84   &  47.16  &  42.06  & 35.96  &  53.20   \\
      & UIE 
      & 74.45    &  66.09   & 58.24   & 53.01   &  48.24  &  45.77  & 48.41  & \underline{56.32} \\
      & AdaKGC 
      & 74.45    &  66.05   & 58.26   & 53.06   &  48.39  &  45.92  & 48.57  & \textbf{56.39}  \\
\hline
\multicolumn{1}{c}{\multirow{3}[2]{*}{\shortstack{\textbf{Relation} \\ (\textbf{NYT}) \\ \textbf{Rel-S F1}}}} 
      & TANL 
      & 90.13   &  83.11  & 76.63   & 73.03   & 67.03   & 63.40   & 47.13  & 71.49    \\
      & UIE 
      & 90.38   &  81.76  & 74.27   & 71.61   & 65.30   & 62.95   & 59.81  & \textbf{72.30}
      \\
      & AdaKGC 
      & 90.18   &  81.65  & 74.13   & 71.48   & 65.10   & 62.75   & 59.77  & \underline{72.15}    \\
\hline
\multicolumn{1}{c}{\multirow{3}[2]{*}{\shortstack{\textbf{Event Trigger} \\ (\textbf{ACE2005}) \\ \textbf{Evt Tri F1}}}} 
      & TEXT2EVENT 
      & 67.10  & 55.81    & 50.38    & 52.99   & 45.82   & 41.39    & 41.65  & 50.73    \\
      & UIE 
      & 70.94  & 60.00    & 57.01    & 62.52   & 60.39   & 54.90    & 53.62  &  \underline{59.91}    \\
      & AdaKGC 
      & 70.57  & 59.75    & 56.50    & 62.31   & 60.57   & 55.49    & 54.32  & \textbf{59.93}     \\
\hline
\multicolumn{1}{c}{\multirow{3}[2]{*}{\shortstack{\textbf{Event Argument} \\ (\textbf{ACE2005}) \\ \textbf{Evt Arg F1}}}}
      & TEXT2EVENT 
      & 49.32  & 37.83    & 33.43  &  35.49  & 30.67   &  27.49  & 27.79    & 34.57    \\
      & UIE 
      & 50.70  & 41.45    & 39.66  &  44.14  & 43.65   &  38.86  & 37.77   & \underline{42.32}   \\
      & AdaKGC 
      & 51.87  & 42.74    & 40.68  &  45.16  & 43.97   &  40.10  & 39.05   & \textbf{43.37}  \\

\bottomrule
\end{tabular}%

}

\caption{
    Vertical schema expansion results in schema-adaptable knowledge graph construction.
}
\label{tab:vertical-experiment}

\end{table*}

%% file: tables/experiment-hybrid.tex
\begin{table*}[h]
\centering
  \resizebox{0.95\textwidth}{!}{
\begin{tabular}{cl|cccc ccc|c}
\toprule
      & \multicolumn{1}{c|}{\textbf{Model}} & \textbf{Iter 1} & \textbf{Iter 2} & \textbf{Iter 3} & \textbf{Iter 4} & \textbf{Iter 5} & \textbf{Iter 6} & \textbf{Iter 7} & \textbf{AVE} \\
\hline
\multicolumn{1}{c}{\multirow{3}[2]{*}{\shortstack{\textbf{Entity} \\ (\textbf{NERD}) \\ \textbf{Ent-F1}}}}
      & TANL
      & 68.37    & 58.50  & 54.51   & 52.46   & 48.67  & 44.13  & 41.51  & 52.59   \\
      & UIE 
      & 69.26    & 59.80  & 55.71   & 53.66   & 49.55  & 46.97  & 48.11  & \underline{54.72}   \\
      & AdaKGC 
      & 69.48    & 59.97  & 55.94   & 53.89   & 49.92  & 47.44  & 48.43  & \textbf{55.01}      \\
\hline
\multicolumn{1}{c}{\multirow{3}[2]{*}{\shortstack{\textbf{Relation} \\ (\textbf{NYT}) \\ \textbf{Rel-S F1}}}} 
      & TANL 
      & 88.67   &  81.59   & 76.48   & 72.22   & 72.35   & 61.76   & 57.87  & 72.99   \\
      & UIE 
      & 90.17   &  83.14   & 77.81   & 71.45   & 72.97   & 65.57   & 63.12  & \textbf{74.89}
      \\
      & AdaKGC 
      & 90.07   &  83.06   & 77.68   & 71.38   & 72.94   & 65.73   & 63.32   & \underline{74.88}   \\
\hline
\multicolumn{1}{c}{\multirow{3}[2]{*}{\shortstack{\textbf{Event Trigger} \\ (\textbf{ACE2005}) \\ \textbf{Evt Tri F1}}}} 
      & TEXT2EVENT 
      & 69.26  & 56.99    & 53.32    & 46.03   & 40.44   & 56.86    & 48.84   &  53.11  \\
      & UIE 
      & 74.69  & 66.34    & 63.12    & 63.21   & 59.86   & 53.18    & 53.26   & \underline{61.95}    \\
      & AdaKGC 
      & 74.84  & 66.99    & 63.28    & 63.07   & 60.94   & 54.72    & 54.80   & \textbf{62.66}     \\
\hline
\multicolumn{1}{c}{\multirow{3}[2]{*}{\shortstack{\textbf{Event Argument} \\ (\textbf{ACE2005}) \\ \textbf{Evt Arg F1}}}}
      & TEXT2EVENT 
      & 50.32  & 38.14    & 35.79  &  31.94  & 28.79   &  37.75  & 33.54   & 36.61   \\
      & UIE 
      & 51.94  & 45.22    & 42.97  &  43.12  & 40.66   &  36.66  & 36.17   & \underline{42.39}   \\
      & AdaKGC 
      & 55.08  & 48.58    & 46.19  &  46.27  & 45.22   & 40.52   & 40.41   & \textbf{46.04} \\

\bottomrule
\end{tabular}%

}

\caption{
    Hybrid schema expansion results in schema-adaptable knowledge graph construction.
}
\label{tab:hybrid-experiment}

\end{table*}

%% file: tables/errorcase-table.tex
\begin{table*}[!t]
\label{tab:study}
\centering
\small
\resizebox{\linewidth}{!}{
\begin{tabular}
{m{1.8cm}|m{8.5cm}|m{2.3cm}|m{2.3cm}|m{1.2cm}}
\toprule
Error Analysis               & Input Example                                                                                    & Gold Type(s)                                                                & Predicted Type(s) 
& Proportion
\\
\hline
\textbf{Weak  \quad  \quad  \quad Transfer}       & Kelly, who declined to talk to reporters here, travel to Tokyo Sunday for talks with Japanese officials.                                   & \color{blue}{Meet[travel]} & \color{red}{Transport[travel]} &{38\%} \\
\hline
\textbf{Inheritance \quad Deficiency}
 & He started his entertainment career at ABC, where he is credited with creating the 'movie of the week' concept.
&\color{blue}{Start-Position[started]} 
& \color{red}{Personnel[started]} &{24\%} \\
\hline
\textbf{Relevance \quad  \quad Neglect}
& Kelly, the US assistant secretary for East Asia and Pacific Affairs, arrived in Seoul from Beijing Friday to brief Yoon, the foreign minister.
&\color{blue}{Transport[arrived]} 
\color{blue}{Meet[brief]} 
& \color{red}{Transport[arrived]} &{21\%} \\
\hline
\textbf{Class \quad  \quad  Imbalance}       & Within weeks he was arrested and charged with sodomising an official driver several years previously and with abusing his powers to cover up the offence.              
&\color{blue}{Arrest-Jail[arrested]} 
\color{blue}{Charge-Indict[charged]}
& \color{red}{Transport[arrested]} &{13\%} \\
\hline
\textbf{Potential \quad \quad Annotation}
 & Anne-Marie will get the couple's 19-room home in New York state, which was on the market last year for 21.5 million dollars, as well as their fine art collection.
&\color{blue}{None Event} 
& \color{red}{Transfer-ownership[get]} 
&\color{black}{4\%} 
\\

\bottomrule
\end{tabular}
}
\caption{
Error analysis on all ACE2005 schema expansion datasets. }
\label{tab:error-analysis}
\end{table*}

%% file: tables/raw_datesets_statistics.tex
\begin{table}[h]
  \centering
  \resizebox{0.49\textwidth}{!}{
    \begin{tabular}{c|ccc|ccc}
    \toprule
           & \#Maj & \#Sub & \#Train & \#Val & \#Test \\
    \midrule
    NERD     & 8     & 66     & 131,767  & 18,824   &  37,648  \\
    NYT       & 4    & 24    & 56,196  & 5,000  & 5,000  \\
    ACE-2005    & 8     & 33    & 19,216  & 901   & 676  \\
    \bottomrule
    \end{tabular}%
  }
  \caption{
    Raw datasets statistics.
    \#Maj indicates the number of major classes, \#Sub is the number of sub-classes, and  \#Val  \#Test is the number of sentences.
}
  \label{tab:details_datasets}
\end{table}%

%% file: tables/schema-adaptable-statistics.tex
\begin{table*}[tbh]
  \centering
  \resizebox{0.80\textwidth}{!}{
    \begin{tabular}{c|ccc|ccc|ccc|ccc}
     \toprule
 \multicolumn{1}{c|}{\multirow{2}{*}{\textbf{Taxonomy}}}& \multicolumn{3}{c|}{\textbf{Horizontal}} & \multicolumn{3}{c|}{\textbf{Vertical }} & \multicolumn{3}{c|}
 {\textbf{Hybrid }} & \multicolumn{3}{c}{\textbf{Analogous}} \\
\cline{2-13}
 & \#Init & \#Add & \#N &
           \#Init & \#Add & \#N&
           \#Init & \#Add & \#N&
           \#Init & \#Add & \#N  \\
    \midrule
    NERD     & 30     & 6   &  7  & 30     & 6   &  7   & 30     & 6   &  7   &66   & -   &  7    \\
    NYT       & 10    & 2   & 8   & 10    & 2   & 8  & 10    & 2   & 8   &24   & -   &  7   \\
    ACE-2005    & 15     & 3  & 7  & 15     & 3  & 7  & 15     & 3  & 7   &33   & -   &  7    \\
    \bottomrule
    \end{tabular}%
  }
  \caption{
    Schema-adaptable datasets statistics.
    \#Init indicates the number of initial subclasses, \#Add is the number of subclasses added per iteration, and \#N is the total number of iterations. 
}
  \label{tab:Horizontal node expansion datasets}
\end{table*}%

%% file: algorithm/alg-horizontal.tex
\begin{algorithm}[t]
	\caption{The construction process of horizontal schema expansion.}  
	\label{alg:horizontal} 
\begin{algorithmic}[1]

\STATE Set sampling seed $\theta$, total iteration $N$,  raw schema $S_{raw}$, and raw dataset $\{D_{train}^{raw},D_{dev}^{raw},D_{test}^{raw}\}$
\STATE Initialize blank schema $S$, blank dataset $\{D_{train},D_{dev},D_{test}\}$
 and initial node number $n_{init}$, node number $n_{iter}$ per iteration

\STATE Randomly select $n_{init}$ nodes in $S_{raw}$, $S$ = $S$ $\cup$ init node, 
$S_1 = S$,$S_{raw} = S_{raw} - S$
\STATE Pick out the instance associated with $S$, $\{D_{train}^1,D_{dev}^1,D_{test}^1\}= \{D_{train},D_{dev},D_{test}\}$

\FOR{iteration $N$}    

\STATE  Calculate $\phi_{Sim}(\vec{W_f}, \vec{W_s})=\frac{\sum_{i=1}^{|W_s|} w_i \cdot w_f}{\|\vec{W_f}\|_2 \cdot\|\vec{W_s}\|_2}$
for node between $S_{raw}$ and $S$

\STATE Pick out top $n_{iter}$ schema node, 
$S$ = $S$ $\cup$ $S_{raw}[:n_{iter}]$, $S_{raw} = S_{raw}- S_{raw}[:n_{iter}]$ 

\STATE Iteration $i$ dataset $S_i = S$,$\{D_{dev}^i,D_{test}^i\} =\{D_{dev},D_{test}\}$

\ENDFOR

\end{algorithmic}
\end{algorithm}

%% file: algorithm/alg-vertical.tex
\begin{algorithm}[t]
	\caption{The construction process of vertical schema expansion.}  
	\label{alg:vertical} 
\begin{algorithmic}[1]

\STATE Set sampling seed $\theta$, total iteration $N$,  raw schema $S_{raw}$, and raw dataset $\{D_{train}^{raw},D_{dev}^{raw},D_{test}^{raw}\}$
\STATE Initialize blank schema $S$, blank dataset $\{D_{train},D_{dev},D_{test}\}$
 and initial node number $n_{init}$, node number $n_{iter}$ per iteration
\FOR{ major node in $S_{raw}$} 
\STATE  $S= S\cup$ major node
\ENDFOR
\STATE Randomly select $n_{init}$ nodes in $S_{raw}$, $S$ = $S$ $\cup$ init node, 
$S_1 = S$,$S_{raw} = S_{raw} - S$
\STATE Pick out the instance associated with $S$, $\{D_{train}^1,D_{dev}^1,D_{test}^1\}= \{D_{train},D_{dev},D_{test}\}$

\FOR{iteration $N$}    

\STATE  Randomly select $n_{iter}$ sub node,
whose parent belongs to $S$
\STATE  $S$ = $S$ $\cup$ $S_{raw}[:n_{iter}]$, $S_{raw} = S_{raw}- S_{raw}[:n_{iter}]$ 

\STATE Iteration $i$ dataset $S_i = S$,$\{D_{dev}^i,D_{test}^i\} =\{D_{dev},D_{test}\}$

\ENDFOR

\end{algorithmic}
\end{algorithm}

%% file: algorithm/alg-hybrid.tex
\begin{algorithm}[t]
	\caption{The construction process of hybrid schema expansion.}  
	\label{alg:hybrid} 
\begin{algorithmic}[1]

\STATE Set sampling seed $\theta$, hybrid ratio $\alpha$,total iteration $N$,  raw schema $S_{raw}$, and raw dataset $\{D_{train}^{raw},D_{dev}^{raw},D_{test}^{raw}\}$
\STATE Initialize blank schema $S$, blank dataset $\{D_{train},D_{dev},D_{test}\}$
 and initial node number $n_{init}$, node number $n_{iter}$ per iteration

\STATE Randomly select $n_{init}$ nodes in $S_{raw}$, $S$ = $S$ $\cup$ init node, 
$S_1 = S$,$S_{raw} = S_{raw} - S$
\STATE Pick out the instance associated with $S$, $\{D_{train}^1,D_{dev}^1,D_{test}^1\}= \{D_{train},D_{dev},D_{test}\}$

\FOR{iteration $N$}    

\IF{random(0,1)< $\alpha$}
\STATE  Calculate $\phi_{Sim}(\vec{W_f}, \vec{W_s})$$=$$\frac{\sum_{i=1}^{|W_s|} w_i \cdot w_f}{\|\vec{W_f}\|_2 \cdot\|\vec{W_s}\|_2}$
for node between $S_{raw}$ and $S$
\STATE Pick out top $n_{iter}$ schema node, 
$S$ = $S$ $\cup$ $S_{raw}[:n_{iter}]$, $S_{raw} = S_{raw}- S_{raw}[:n_{iter}]$ 

\ELSE

\STATE  Randomly select $n_{iter}$ sub node,
whose parent belongs to $S$
\STATE  $S$ = $S$ $\cup$ $S_{raw}[:n_{iter}]$, $S_{raw} = S_{raw}- S_{raw}[:n_{iter}]$ 

\ENDIF

\STATE Iteration $i$ dataset $S_i = S$,$\{D_{dev}^i,D_{test}^i\} =\{D_{dev},D_{test}\}$

\ENDFOR

\end{algorithmic}
\end{algorithm}

%% file: algorithm/alg-analogous.tex
\begin{algorithm}[t]
	\caption{The construction process of analogous schema expansion.}  
	\label{alg:analougous} 
\begin{algorithmic}[1]

\STATE Set sampling seed $\theta$, total iteration $N$,  raw schema $S_{raw}$, and raw dataset $\{D_{train}^{raw},D_{dev}^{raw},D_{test}^{raw}\}$
\STATE Initialize blank schema $S$, blank dataset $\{D_{train},D_{dev},D_{test}\}$
 and  node number $n_{iter}$ per iteration

\STATE Initialize $S=S_{raw}$, $S_1 = S$,
$\{D_{train},D_{dev},D_{test}\}$$=$$\{D_{train}^{raw},D_{dev}^{raw},D_{test}^{raw}\}$

\STATE Pick out the instance associated with $S$, $\{D_{train}^1,D_{dev}^1,D_{test}^1\}= \{D_{train},D_{dev},D_{test}\}$

\FOR{iteration $N$}    

\STATE Randomly select $n_{iter}$ schema node in $S$,
calculate $\vec{v}(W)=\left\{\sum_{w_r \in W} NMPI\left(w_r, w_j\right)^\gamma\right\}_{j=1, \ldots,|W_L|}$
for each node, create candidate nodes by pairing $W_r$ with all words $W_L$ in the corpus word list

\STATE Replace $S[:n_{iter}]$ with candidate nodes with analogous semantics

\STATE Iteration $i$ dataset $S_i = S$,$\{D_{dev}^i,D_{test}^i\} =\{D_{dev},D_{test}\}$

\ENDFOR

\end{algorithmic}
\end{algorithm}

%% file: tables/experiment-analogous.tex
\begin{table*}[b]
\centering
  \resizebox{0.85\textwidth}{!}{
\begin{tabular}{cl|cccc ccc|c}
\toprule
      & \multicolumn{1}{c|}{\textbf{Model}} & \textbf{Iter 1} & \textbf{Iter 2} & \textbf{Iter 3} & \textbf{Iter 4} & \textbf{Iter 5} & \textbf{Iter 6} & \textbf{Iter 7} & \textbf{AVE} \\
\hline
\multicolumn{1}{c}{\multirow{3}[2]{*}{\shortstack{\textbf{Entity} \\ (\textbf{NERD}) \\ \textbf{Ent-F1}}}} 
      & TANL
      & 68.34  & 62.70    & 57.22   & 52.94   & 47.73   & 42.24   & 36.09   &  52.47  \\
      & UIE 
      & 67.58  & 63.39    & 60.37   & 58.00   &  58.17  &  54.26   & 49.53   & \underline{58.76}   \\
      & AdaKGC 
      & 68.22  & 63.87    & 61.01   & 58.18  &  58.76   &  54.58   & 49.81    &   \textbf{59.20}  \\
\hline
\multicolumn{1}{c}{\multirow{3}[2]{*}{\shortstack{\textbf{Relation} \\ (\textbf{NYT}) \\ \textbf{Rel-S F1}}}} 
      & TANL 
      & 89.80  & 86.82    & 83.51   &  77.99  &  73.61  & 73.32   &  66.37  &  \textbf{78.77}  \\
      & UIE 
      & 89.66  & 86.56    & 83.27   &  77.80  & 73.30   &  73.27  & 66.33   &  \underline{78.60}   \\
      & AdaKGC 
      & 89.17  & 86.12    & 82.76   &  77.33  & 72.85   & 72.91   & 65.84   &  78.14   \\
\hline
\multicolumn{1}{c}{\multirow{3}[2]{*}{\shortstack{\textbf{Event Trigger} \\ (\textbf{ACE2005}) \\ \textbf{Evt Tri F1}}}} 
      & TEXT2EVENT 
      & 64.40   &  60.39   & 53.96  & 52.64  & 40.83  &  32.45  & 29.58   & 47.75   \\
      & UIE 
      & 69.69   &  64.53   & 62.10  & 58.47  & 49.76  &  42.51  & 39.31   &  \underline{55.20}   \\
      & AdaKGC 
      & 69.63   &  64.77   & 61.86  & 58.31  & 50.08  & 43.07   & 40.40   &  \textbf{55.45}   \\
\hline
\multicolumn{1}{c}{\multirow{3}[2]{*}{\shortstack{\textbf{Event Argument} \\ (\textbf{ACE2005}) \\ \textbf{Evt Arg F1}}}}
      & TEXT2EVENT 
      &  45.88  & 40.96   & 34.21  & 33.40   & 24.49  & 19.62  & 16.77   & 30.76    \\
      & UIE 
      & 49.96   & 42.67   & 40.50  & 36.56   & 28.70  & 24.06  & 20.47   & \underline{34.70}   \\
      & AdaKGC 
      & 51.74   & 44.56   & 42.48  & 32.61   & 30.52  & 25.41  & 22.10   & \textbf{35.63}  \\
\bottomrule
\end{tabular}%

}
\caption{
    Analogous schema replacement results in schema-adaptable knowledge graph construction. 
}
\label{tab:Analogous-experiment}

\end{table*}

%% file: tables/data-example-hor-ace.tex
\begin{table*}[tbh]
\centering
\resizebox{0.99\textwidth}{!}{

    \begin{tabular}{p{.8\textwidth} p{.2\textwidth}}
\toprule
\textit{Input:}   The Belgrade district court said that Markovic will be tried along with 10 other Milosevic-era officials who face similar charges of ‘inappropriate use of state property’ that carry a sentence of up to five years in jail.                      
& Labels \\
 \midrule
\textit{ Iteration 1 schema:} "attack", "start position", "transfer ownership", "be born", "sentence", "die", "arrest jail", "transport", "elect", "phone write", "end organization", "sue", "acquit", "marry", "extradite"
  &  Sentence[sentence]
  \\
 \midrule
\textit{ Iteration 2 schema:} "attack", "start position", "transfer ownership", "be born", "sentence", "die", "arrest jail", "transport", "elect", \textcolor{blue}{"injure"}, "phone write", \textcolor{blue}{"fine"}, \textcolor{blue}{"convict"}, "end organization", "sue", "acquit", "marry", "extradite"
  &  Sentence[sentence] 
  \\
 \midrule
\textit{ Iteration 3 schema:} "attack", "start position", "transfer money", \textcolor{blue}{"transfer ownership"}, "be born", "sentence", "die", \textcolor{blue}{"demonstrate"}, "arrest jail", "transport", "elect", "injure", "phone write", "fine", "convict", "end organization", "sue", "acquit", \textcolor{blue}{"execute"}, "marry", "extradite"
  &   Sentence[sentence]
  \\
   \midrule
\textit{ Iteration 4 schema:} \textcolor{blue}{"end position"}, "attack", "start position", "transfer money", "transfer ownership", "be born", "sentence", "die", "demonstrate", "arrest jail", "transport", "elect", \textcolor{blue}{"start organization"}, "injure", "phone write", "fine", "convict", "end organization", "sue", "acquit", "execute", "marry", "extradite", \textcolor{blue}{"pardon"}
  &   Sentence[sentence]
  \\
 \midrule
\textit{ Iteration 5 schema:} "end position", "attack", "start position", "transfer money", "transfer ownership", "be born", "sentence", "die", "demonstrate", "arrest jail", "transport", "elect", "start organization", "injure", "phone write", \textcolor{blue}{"declare bankruptcy"}, \textcolor{blue}{"trial hearing"}, "fine", "convict", "end organization", "sue", "acquit", \textcolor{blue}{"appeal"}, "execute", "marry", "extradite", "pardon"
  &   Sentence[sentence] \color{purple}{Trial hearing[tried]}
  \\
 \midrule
\textit{ Iteration 6 schema:} "end position", "attack", "start position", \textcolor{blue}{"charge indict"}, "transfer money", "transfer ownership", \textcolor{blue}{"release parole"}, "be born", "sentence", "die", "demonstrate", "arrest jail", "transport", "elect", "start organization", "injure", "phone write", \textcolor{blue}{"merge organization"}, "declare bankruptcy", "trial hearing", "fine", "convict", "end organization", "sue", "acquit", "appeal", "execute", "marry", "extradite", "pardon"
  &   Sentence[sentence] Trial hearing[tried] \quad \quad \color{purple}{Charge-Indict[charges]}
  \\
 \midrule
\textit{ Iteration 7 schema:} "end position", "attack", "start position", \textcolor{blue}{"nominate"}, "charge indict", "transfer money", "transfer ownership", "release parole", "be born", "sentence", "die", "demonstrate", "arrest jail", "transport", "elect", "start organization", \textcolor{blue}{"meet"}, "injure", "phone write", "merge organization", "declare bankruptcy", "trial hearing", "fine", "convict", "end organization", "sue", \textcolor{blue}{"divorce"}, "acquit", "appeal", "execute", "marry", "extradite", "pardon"
  &   Sentence[sentence] Trial hearing[tried] \quad \quad Charge-Indict[charges] 
  \\

\bottomrule
\end{tabular}
}
\caption{Adaptive evolution of horizontal schema expansion on ACE2005 dataset.}
\label{tab:dataset-horizontal-ace}
\end{table*}

%% file: tables/data-example-ver-ace.tex
\begin{table*}[tbh]
\centering
\resizebox{0.99\textwidth}{!}{

    \begin{tabular}{p{.8\textwidth} p{.2\textwidth}}
\toprule
\textit{Input:}  Kelly, the US assistant secretary for East Asia and Pacific Affairs, arrived in Seoul from Beijing Friday to brief Yoon, the foreign minister.
& Labels \\
 \midrule
\textit{ Iteration 1 schema:} \underline{"personnel"}, "attack", \underline{"justice"}, "transfer money", "transfer ownership", "release parole", "be born", "sentence", "die", "demonstrate", "transport", \underline{"business"}, \underline{"contact"}, \underline{"life"}, "fine", "sue", "execute", "marry", "extradite", "pardon"
  & Transport[arrived] Contact[brief]
  \\
 \midrule
\textit{ Iteration 2 schema:} \underline{"personnel"}, "attack", \underline{"justice"}, "transfer money", "transfer ownership", "release parole", "be born", "sentence", "die", "demonstrate", "transport",\underline{"business"}, \textcolor{blue}{"meet"}, \underline{"life"}, \underline{"contact"}, "fine", "sue", \textcolor{blue}{"acquit"}, \textcolor{blue}{"appeal"}, "execute", "marry", "extradite", "pardon"
  & Transport[arrived] \textcolor{purple}{Meet[brief]}
  \\
 \midrule
\textit{ Iteration 3 schema:} \underline{"personnel"}, "attack", \underline{"justice"}, "transfer money", "transfer ownership", "release parole", "be born", "sentence", "die", "demonstrate", "transport", \textcolor{blue}{"start organization"}, "meet", \underline{"life"}, \underline{"contact"},\textcolor{blue}{ "merge organization"}, \underline{"business"}, \textcolor{blue}{"trial hearing"}, "fine", "sue", \underline{"acquit"}, \underline{"appeal"}, "execute", "marry", "extradite", "pardon"
  &  Transport[arrived] Meet[brief]
  \\
 \midrule
\textit{ Iteration 4 schema:} \underline{"personnel"}, "attack", \underline{"justice"}, "transfer money", "transfer ownership", "release parole", "be born", "sentence", "die", "demonstrate", "transport", "start organization", "meet", \underline{"life"}, \textcolor{blue}{"phone write"}, "merge organization", \textcolor{blue}{"declare bankruptcy"}, "trial hearing", "fine", \textcolor{blue}{"end organization"}, "sue", "acquit", "appeal", "execute", "marry", "extradite", "pardon"
  &  Transport[arrived] Meet[brief]
  \\
 \midrule
\textit{ Iteration 5 schema:} \underline{"personnel"}, "attack",  \textcolor{blue}{"charge indict"}, "transfer money", "transfer ownership", "release parole", "be born", "sentence", "die", "demonstrate",  \textcolor{blue}{"arrest jail"}, "transport", "start organization", "meet", \underline{"life"}, "phone write", "merge organization", "declare bankruptcy", "trial hearing", "fine",  \textcolor{blue}{"convict"}, "end organization", "sue", "acquit", "appeal", "execute", "marry", "extradite", "pardon"
  &  Transport[arrived] Meet[brief]
  \\
 \midrule
\textit{ Iteration 6 schema:}  \textcolor{blue}{"end position"}, "attack", \underline{"personnel"}, "charge indict", "transfer money", "transfer ownership", "release parole", "be born", "sentence", "die", "demonstrate", "arrest jail", "transport", "start organization", "meet",  \textcolor{blue}{"injure"}, "phone write", "merge organization", "declare bankruptcy", "trial hearing", "fine", "convict", "end organization", "sue", "divorce", "acquit",  \textcolor{blue}{"appeal"}, "execute", "marry", "extradite", "pardon"
  &  Transport[arrived] Meet[brief]
  \\
 \midrule
\textit{ Iteration 7 schema:} "end position", "attack",  \textcolor{blue}{"start position"},  \textcolor{blue}{"nominate"}, "charge indict", "transfer money", "transfer ownership", "release parole", "be born", "sentence", "die", "demonstrate", "arrest jail", "transport",  \textcolor{blue}{"elect"}, "start organization", "meet", "injure", "phone write", "merge organization", "declare bankruptcy", "trial hearing", "fine", "convict", "end organization", "sue", "divorce", "acquit", "appeal", "execute", "marry", "extradite", "pardon"
  &  Transport[arrived] Meet[brief]
  \\

\bottomrule
\end{tabular}
}
\caption{Adaptive evolution of vertical schema expansion on ACE2005 dataset. Underlined classes refer to major classes, which will be covered by refined sub classes. }
\label{tab:dataset-vertical-ace}
\end{table*}

%% file: tables/data-example-mix-ace.tex
\begin{table*}[tbh]
\centering
\resizebox{0.99\textwidth}{!}{

    \begin{tabular}{p{.8\textwidth} p{.2\textwidth}}
\toprule
\textit{Input:}  The charismatic leader of Turkey's governing party was named prime minister Tuesday, a step that probably boosts chances that the United States will get permission to deploy troops in the country along Iraq's northern border.                        
& Labels \\
 \midrule
\textit{ Iteration 1 schema:} "attack", "justice", "transfer money", "transfer ownership", "release parole", "be born", "sentence", "die", "demonstrate", "transport", "life", "fine", "sue", "execute", "marry", "extradite", "pardon" 
  &  Transport[deploy] 
  \\
 \midrule
 \textit{Iteration 2 schema:} "attack", "justice", "transfer money", "transfer ownership", "release parole", "be born", "sentence", "die", "demonstrate", "transport", 
\textcolor{blue}{"meet"}, "life", 
\textcolor{blue}{\underline{"contact"}}, "fine", "sue", \textcolor{blue}{"acquit"}, \textcolor{blue}{"appeal"}, "execute", "marry", "extradite", "pardon" &  Transport[deploy]   \\
 \midrule
 \textit{Iteration 3 schema:} "attack", "justice", "transfer money", "transfer ownership", "release parole", "be born", "sentence", "die", "demonstrate", "transport", \textcolor{blue}{"start organization"}, "meet", "life", "contact", \textcolor{blue}{"merge organization"}, \textcolor{blue}{\underline{"business"}}, \textcolor{blue}{"trial hearing"}, "fine", "sue", "acquit", "appeal", "execute", "marry", "extradite", "pardon" &  Transport[deploy] \\
 \midrule
 \textit{Iteration 4 schema: }"attack", "justice", "transfer money", "transfer ownership", "release parole", "be born", "sentence", "die", "demonstrate", "transport", "start organization", "meet", "life", \textcolor{blue}{"phone write"}, "merge organization", \textcolor{blue}{"declare bankruptcy"}, "trial hearing", "fine", \textcolor{blue}{"end organization"}, "sue", "acquit", "appeal", "execute", "marry", "extradite", "pardon"  &  Transport[deploy]   \\
  \midrule
  \textit{Iteration 5 schema: }"attack", \textcolor{blue}{"charge indict"}, "transfer money", "transfer ownership", "release parole", "be born", "sentence", "die", "demonstrate", \textcolor{blue}{"arrest jail"}, "transport", "start organization", "meet", "life", "phone write", "merge organization", "declare bankruptcy", "trial hearing", "fine", \textcolor{blue}{"convict"}, "end organization", "sue", "acquit", "appeal", "execute", "marry", "extradite", "pardon"  &  Transport[deploy]  \\
    \midrule
\textit{Iteration 6 schema: } \textcolor{blue}{"end position"}, "attack", \textcolor{blue}{\underline{"personnel"}}, "charge indict", "transfer money", "transfer ownership", "release parole", "be born", "sentence", "die", "demonstrate", "arrest jail", "transport", "start organization", "meet", \textcolor{blue}{"injure"}, "phone write", "merge organization", "declare bankruptcy", "trial hearing", "fine", "convict", "end organization", "sue",  \textcolor{blue}{"divorce"}, "acquit", "appeal", "execute", "marry", "extradite", "pardon"  & Transport[deploy] \color{purple}{Personnel[named]}
  \\
\midrule
\textit{Iteration 7 schema: } "end position", "attack", \textcolor{blue}{"start position"}, \textcolor{blue}{"nominate"}, "charge indict", "transfer money", "transfer ownership", "release parole", "be born", "sentence", "die", "demonstrate", "arrest jail", "transport", \textcolor{blue}{"elect"}, "start organization", "meet", "injure", "phone write", "merge organization", "declare bankruptcy", "trial hearing", "fine", "convict", "end organization", "sue", "divorce", "acquit", "appeal", "execute", "marry", "extradite", "pardon"
  &  Transport[deploy]  \color{purple}{Elect[named]}
  \\
\bottomrule
\end{tabular}
}
\caption{Adaptive evolution of hybrid schema expansion on ACE2005 dataset. Underlined classes refer to father classes, which occurs when directly adding sub classes that  corresponding major class not exists. }
\label{tab:dataset-hybrid-ace}
\end{table*}

%% file: tables/data-example-ana-ace.tex
\begin{table*}[tbh]
\centering
\resizebox{0.99\textwidth}{!}{

    \begin{tabular}{p{.8\textwidth} p{.2\textwidth}}
\toprule
\textit{Input:}  Webb also said details of the breakdowns of the Welches' previous marriages were likely to come up , and cited reports of alleged extramarital affairs by both.         
& Labels \\
 \midrule
\textit{ Iteration 1 schema:} "end position", "attack", "start position", "nominate", "charge indict", "transfer money", "transfer ownership", "release parole", "be born", "sentence", "die", "demonstrate", "arrest jail", "transport", "elect", "start organization", "meet", "injure", "phone write", "merge organization", "declare bankruptcy", "trial hearing", "fine", "convict", "end organization", "sue", "divorce", "acquit", "appeal", "execute", "marry", "extradite", "pardon"
  & Divorce[breakdowns] Marry[marriages]
  \\
 \midrule
\textit{ Iteration 2 schema:}  "end position", "attack", \textcolor{blue}{"begin"}, "nominate", "charge indict", "transfer money", "transfer ownership", "release parole", "be born", "sentence", "die", "demonstrate", "arrest jail", \textcolor{blue}{"carry"}, "elect", "start organization", "meet", "injure", "phone write", "merge organization", "declare bankruptcy", "trial hearing", "fine", "convict", "end organization", "sue", \textcolor{blue}{"separate"}, "acquit", "appeal", "execute", "marry", "extradite", "pardon"
  & \textcolor{purple}{Separate[breakdowns]} Marry[marriages]
  \\
 \midrule
\textit{ Iteration 3 schema:}  \textcolor{blue}{"end"}, "attack", "begin", "nominate", \textcolor{blue}{"prosecute"}, \textcolor{blue}{"remittance"}, "transfer ownership", "release parole", "be born", "sentence", "die", "demonstrate", "arrest jail", "carry", "elect", "start organization", "meet", "injure", "phone write", "merge organization", "declare bankruptcy", "trial hearing", "fine", "convict", "end organization", "sue", "separate", "acquit", "appeal", "execute", "marry", "extradite", "pardon"
  &  Separate[breakdowns] Marry[marriages]
  \\
   \midrule
\textit{ Iteration 4 schema:} "end", "attack", "begin", "nominate", "prosecute", "remittance", "transfer ownership", "release parole", "be born", "sentence",  \textcolor{blue}{"pass away"}, "demonstrate", "arrest jail", "carry", "elect", "start organization", "meet", "injure", "phone write", "merge organization", "declare bankruptcy",  \textcolor{blue}{"attend the trial"}, "fine", "convict", "end organization", "sue", "separate", "acquit", "appeal",  \textcolor{blue}{"perform"}, "marry", "extradite", "pardon"
  &  Separate[breakdowns] Marry[marriages]
  \\
 \midrule
\textit{ Iteration 5 schema:} "end", "attack", "begin", "nominate", "prosecute", "remittance", "transfer ownership", "release parole", "be born",  \textcolor{blue}{"condemn"}, "pass away", "demonstrate", "arrest jail", "carry", "elect", "start organization",  \textcolor{blue}{"encounter"}, "injure", "phone write", "merge organization",  \textcolor{blue}{"go out of business"}, "attend the trial", "fine", "convict", "end organization", "sue", "separate", "acquit", "appeal", "perform", "marry", "extradite", "pardon"
  &   Separate[breakdowns] Marry[marriages]
  \\
 \midrule
\textit{ Iteration 6 schema:} "end", "attack", "begin", "nominate", "prosecute", "remittance",  \textcolor{blue}{"giveaway"}, "release parole", "be born", "condemn", "pass away",  \textcolor{blue}{"parade"}, "arrest jail", "carry",  \textcolor{blue}{"vote"}, "start organization", "encounter", "injure", "phone write", "merge organization", "go out of business", "attend the trial", "fine", "convict", "end organization", "sue", "separate", "acquit", "appeal", "perform", "marry", "extradite", "pardon"
  &  Separate[breakdowns] Marry[marriages]
  \\
 \midrule
\textit{ Iteration 7 schema:} "end", "attack", "begin", "nominate", "prosecute", "remittance", "giveaway", "release parole", "be born", "condemn", "pass away", "parade", "arrest jail", "carry", "vote", "start organization", "encounter", \textcolor{blue}{"hurt"}, \textcolor{blue}{"communication"}, "merge organization", "go out of business", "attend the trial", "fine", "convict", "end organization", "sue", "separate", "acquit", "appeal", "perform", \textcolor{blue}{"wed"}, "extradite", "pardon"
  &  Separate[breakdowns] \textcolor{purple}{Wed[marriages]}
  \\

\bottomrule
\end{tabular}
}
\caption{Adaptive evolution of analogous schema expansion on ACE2005 dataset.}
\label{tab:dataset-ana-ace}
\end{table*}

%% file: tables/gpt3-5.tex
\begin{table*}[tbh]
\centering
\resizebox{0.99\textwidth}{!}{

\begin{tabular}{p{.9\textwidth}}
\toprule
\textit{GPT-3.5 Input Example:}         \\
 \midrule
\textcolor{blue}{There are some relation extraction samples, relation must be taken from schema, head entity and tail entity must be taken from context. Relation, head entity and tail entity may have multiple.}

\textcolor{violet}{schema: ["people", "country", "religion", "major shareholder of", "industry", "contains", "brith place", "location", "nationality", "advisors", "neighborhood of", "place lived", "capital", "geographic distribution", "teams", "major shareholders", "place of death", "children", "company", "profession", "place founded", "founders"]}

Context: In Queens, North Shore Towers, near the Nassau border, supplanted a golf course, and housing replaced a gravel quarry in Douglaston.

The relation involved in the above sentence are: 1. The head entity is Douglaston, relation is neighborhood of, tail entity is Queens; 2. The head entity is Queens, relation is contains, tail entity is Douglaston.



Context: Martin, the district attorney for Lehigh County in Pennsylvania, said that after his office's review of the records, he was satisfied with Mr. Cullen's denials.

The relation involved in the above sentence are: 1. The head entity is Pennsylvania, relation is contains, tail entity is Lehigh County.

Context: Mr.Brown has demeaned Mr.Bush as "a cheerleader," declared that Homeland Security Secretary Michael Chertoff did not know "the first thing about running a disaster," and called critics like Representative Gene Taylor, Democrat of Mississippi, "a little twerp" and Senator Norm Coleman, Republican of Minnesota, an unprintable vulgarity (both in Playboy).

The relation involved in the above sentence are: 1. The head entity is Gene Taylor, relation is place lived, tail entity is Mississippi.

...

\textcolor{blue}{Do you understand how to do relation extraction based on schema?
Now it's your turn to do relation extraction.}

\textcolor{violet}{schema: ["people", "country", "religion", "major shareholder of", "industry", "contains", "birth place", "location", "nationality", "advisors", "neighborhood of", "place lived", "capital", "geographic distribution", "teams", "major shareholders", "place of death", "children", "company", "profession", "place founded", "founders"]}

Context: But that spasm of irritation by a master intimidator was minor compared with what Bobby Fischer, the erratic former world chess champion, dished out in March at a news conference in Reykjavik, Iceland.

\textcolor{red}{The relation involved in the above sentence are:}
  \\
\midrule
\textit{GPT-3.5 Output Example:} 
The relation involved in the above sentence are:  
1. The head entity is Bobby Fischer, relation is place lived, tail entity is Iceland; 
2. The head entity is Iceland, relation is contains, tail entity is Reykjavik;
3. The head entity is Iceland, relation is capital, tail entity is Reykjavik.
  \\
\midrule
\textit{Golden Output Example:} 
The relation involved in the above sentence are:  1. The head entity is Bobby Fischer, relation is nationality, tail entity is Iceland;
2. The head entity is Iceland, relation is capital, tail entity is Reykjavik;
3. The head entity is Iceland, relation is contains, tail entity is Reykjavik;
4. The head entity is Bobby Fischer, relation is place of death, tail entity is Reykjavik. 
\\

\bottomrule
\end{tabular}
}
\caption{Examples of GPT-3.5 experiment on NYT dataset. A total of 20 demonstrations are given to the model.}
\label{tab:gpt3.5-input}
\end{table*}